\documentclass[accepted]{article}

\usepackage{times}
\usepackage{graphicx} 
\usepackage{subfigure} 

\usepackage{natbib}

\usepackage{algorithm}
\usepackage{algorithmic}

\usepackage{amsmath}
\usepackage{amsfonts}
\usepackage{amssymb}
\usepackage{graphicx}
\usepackage{bm}

\usepackage{hyperref}

\usepackage[accepted]{icml2017} 

\newtheorem{remark}{Remark}
\newtheorem{corollary}{Corollary}

\newcommand{\norm}[1]{{\|{#1}\|}}

\renewcommand{\v}[2]{\ensuremath{v_{#2}^{(#1)}}}

\newcommand{\E}{\ensuremath{\mathbb{E}}}

\newcommand{\field}[1]{\mathbb{#1}}

\newcommand{\up}{\mbox{\sc up}}

\newcommand{\bb}{\boldsymbol{b}}

\newcommand{\bz}{\boldsymbol{z}}
\newcommand{\bx}{\boldsymbol{x}}

\newcommand{\bw}{\boldsymbol{w}}
\newcommand{\bu}{\boldsymbol{u}}

\newcommand{\bv}{\boldsymbol{v}}

\newcommand{\be}{\boldsymbol{e}}
\newcommand{\bzero}{\boldsymbol{0}}

\newcommand{\R}{\field{R}}

\newcommand{\scO}{\mathcal{O}}

\newcommand{\scU}{\mathcal{U}}

\renewcommand{\tilde}{\widetilde}

\renewcommand{\Pr}{\mathbb{P}}

\renewcommand{\v}[2]{\ensuremath{\bv_{#2}^{(#1)}}}

\newcommand{\ignore}[1]{}

\DeclareMathOperator*{\argmax}{argmax}

\newcommand{\CB}{\mbox{\sc cb}}
\newcommand{\HD}{\mbox{\sc hd}}

\newcommand{\hN}{\widehat {N}}

\newtheorem{theorem}{Theorem}
\newtheorem{lemma}{Lemma}

\icmltitlerunning{On Context-Dependent Clustering of Bandits}

\begin{document} 

\twocolumn[
\icmltitle{On Context-Dependent Clustering of Bandits}




\begin{icmlauthorlist}
\icmlauthor{Claudio Gentile}{a}
\icmlauthor{Shuai Li}{b}
\icmlauthor{Purushottam Kar}{d}
\icmlauthor{Alexandros Karatzoglou}{c}
\icmlauthor{Evans Etrue}{a}
\icmlauthor{Giovanni Zappella}{e}

\end{icmlauthorlist}

\icmlaffiliation{a}{DiSTA, University of Insubria, Italy}
\icmlaffiliation{b}{University of Cambridge, United Kingdom}
\icmlaffiliation{c}{Telefonica Research, Spain}
\icmlaffiliation{d}{IIT Kanpur, India}
\icmlaffiliation{e}{Amazon Berlin, Germany}

\icmlcorrespondingauthor{Claudio Gentile}{claudio.gentile@uninsubria.it}


\vskip 0.3in
]



\printAffiliationsAndNotice{}  

\begin{abstract} 
We investigate a novel cluster-of-bandit algorithm CAB for collaborative recommendation tasks that implements the underlying feedback sharing mechanism by estimating 
the neighborhood of users in a context-dependent manner. CAB makes sharp departures from the state of the art by incorporating collaborative effects into inference as well as learning processes in a manner that seamlessly interleaving explore-exploit tradeoffs and collaborative steps. We prove regret bounds under various assumptions on the data, which exhibit a crisp dependence on the expected number of clusters over the users, a natural measure of the statistical difficulty of the learning task. Experiments on production and real-world datasets show that CAB offers significantly increased prediction performance against a representative pool of state-of-the-art methods.
\end{abstract}

\section{Introduction}
%
In many prominent applications of bandit algorithms, such as computational advertising, web-page content optimization and recommendation systems, one of the main sources of information is embedded in the preference relationships between users and the items served. Preference patterns, emerging from clicks, views or purchase of items, are typically exploited through collaborative filtering techniques.

In fact, it is common knowledge in recommendation systems practice that collaborative effects carry more information about user preferences than, say, demographic metadata~(e.g., \cite{pilaszy09}). Yet, as content recommendation functionalities are incorporated in very diverse online services,
the requirements often differ vastly. For instance, in a movie recommendation system, where the catalog is relatively static and ratings for items will accumulate, one can easily deploy collaborative filtering methods such as matrix factorization or restricted Boltzmann machines. The same methods become practically impossible to use in more dynamic environments such as in news or YouTube video recommendation, where we have to deal with a continuous stream of new items to be recommended, along with new users to be served.
%
These dynamic environments pose a dual challenge to recommendation methods:
1) How to present the new items to the users (or, vice versa, which items to present to new users), in order to optimally gather preference information on the new content (exploration), and
2) How to use all the available user-item preference information gathered so far (exploitation). Ideally, one would like to exploit both the content information but also, and more importantly, the collaborative effects that can be observed across users and items.

When the users to serve are many and the content universe (or content popularity) changes
rapidly over time, recommendation services have to show both strong adaptation in matching
user preferences and high algorithmic scalability/responsiveness so as to allow
an effective on-line deployment.
In typical scenarios like social networks, where users are engaged in technology-mediated
interactions influencing each other's behavior, it is often possible to single out a few groups or
{\em communities} made up of users sharing similar interests and/or behavior.
Such communities are not static over time and, more often than not, are clustered around specific
content {\em types}, so that a given set of users can in fact host a multiplex of interdependent
communities depending on specific content items, which can be changing dramatically on the fly.
We call this multiplex of interdependent clusterings over users induced by the content universe
a {\em context-dependent} clustering. In addition to the above, the set of users itself can change over time, for new users get targeted by the service, others may sign out or unregister. Thus, a recommendation method has to readily adapt to 
a changing set of both users and items.


In this paper, we introduce and analyze the CAB (Context-Aware clustering of Bandit) algorithm, a simple and flexible algorithm rooted in the linear contextual bandit framework that does the above by incorporating collaborative effects which traditional approaches to contextual bandits ignore (e.g., \cite{Aue02,lcls10,chu2011contextual,abbasi2011improved}). CAB adapts to match user preferences in the face of a constantly evolving content universe and set of targeted users.
CAB implements the context-dependent clustering intuition by computing clusterings of bandits which allows each content item to cluster users into groups (which are few relative to the total number of users), where within each group, users tend to react similarly when that item gets recommended. CAB distinguishes itself in allowing distinct items to induce distinct clusterings, which is frequently observed in practice (e.g., \cite{sst09}). These clusterings are in turn  suggestive of a natural context-dependent feedback sharing mechanism across users. CAB is thus able to exploit collaborative effects in contextual bandit settings in a manner similar to the way neighborhood techniques are used by batch collaborative filtering.

We analyze our algorithm from both the theoretical and the experimental standpoint. On the theoretical side, we provide a regret analysis where the number of users engaged essentially enters in the regret bound only through the {\em expected} number of context-dependent clusters over the users, a natural measure of the predictive hardness of learning these users. We also extend this result to provide a sharper bound under sparsity assumptions on the user model vectors. On the experimental side, we present comparative evidence on production and real-world datasets that our algorithm significantly outperforms, in terms of prediction performance, state-of-the-art contextual bandit algorithms that either do not leverage any clustering at all or do so in a context-{\em in}dependent fashion.

\subsection{Related Work}
%
The literature on contextual bandit algorithms is too large to be surveyed here. In the sequel, we briefly mention what we believe are the works most closely related to ours. The technique of sequentially clustering users in the bandit setting was introduced in 
\cite{mm14,glz14}, but has also been inspired by earlier references, e.g., \cite{a+13} on transfer learning for stochastic bandits, and
\cite{dkc13} on low-rank (Gaussian Process) bandits. This led to further developments such as \cite{nl14}, which relies on $k$-means clustering, and \cite{ksl16} which proposes distributed clustering of confidence ball algorithms for solving linear bandit problems in peer to peer networks. Related papers that implement feedback sharing mechanisms by leveraging (additional) social information among users include \cite{cgz13,w+16}. In all these cases, the way users are grouped is not context-dependent. Even more related to our work is the recent paper \cite{lkg16} which proposes to simultaneously cluster users as well as items, with item clusters dictating user clusters. However, a severe limitation of this approach is that the content universe has to be finite and known in advance, and in addition to that the resulting algorithm is somewhat involved. Compared to all these previous works, our approach distinguishes itself for being simple and flexible (e.g., we can seamlessly accomodate the inclusion/exclusion of users), as well as for performing feedback propagation among users in a context-dependent manner. As will be demostrated in Section \ref{s:exp}, this offers significant performance boosts in real-world recommendation settings. 



\section{Notation and Preliminaries}
\label{s:prel}
%
We will consider the bandit clustering model standard in the literature, but with the crucial difference that we will allow user behavior similarity to be represented by a family of clusterings that depend on the specific feature (or {\em context}) vector $\bx$ under consideration. In particular, we let $\scU = \{1,\ldots, n\}$ represent the set of $n$ users. An item, represented by its feature vector $\bx \in \R^d$ can be seen as inducing a (potentially different) partition of the user set $\scU$ into a small number $m(\bx)$ of clusters $\{U_1(\bx), U_2(\bx), \ldots, U_{m(\bx)}(\bx)\}$, where $m(\bx) \ll n$. Users belonging to the same cluster $U_j(\bx)$ share similar behavior w.r.t. $\bx$ (e.g., they both like or both dislike the item represented by $\bx$), while users lying in different clusters have significantly different behavior.

This is a much more flexible model that allows users to agree on their opinion of certain items and disagree on others, something that often holds in practice. It is important to note that the mapping $\bx \rightarrow \{U_1(\bx), U_2(\bx), \ldots, U_{m(\bx)}(\bx)\}$ specifying the actual partitioning of $\scU$ into the clusters determined by $\bx$ (including the number of clusters $m(\bx)$), and the common user behavior within each cluster are {\em unknown} to the learner, and have to be inferred based on user feedback.

To make things simple, we assume that the context-dependent clustering is determined by the linear functions $\bx \rightarrow \bu_i^\top\bx$, each one parameterized by an unknown vector $\bu_i \in \R^d$ hosted at user $i \in \scU$, with $||\bu_i|| = 1$ for all $i$, in such a way that if users $i, i' \in \scU$ are in the same cluster w.r.t. $\bx$ then  $\bu_i^\top\bx = \bu_{i'}^\top\bx$, and if $i, i' \in \scU$ are in different clusters w.r.t. $\bx$ then $|\bu_i^\top\bx - \bu_{i'}^\top\bx| \geq \gamma$, for some \emph{gap parameter} $\gamma >0$.\footnote
{
As usual, this hypothesis may be relaxed by assuming the existence
of two thresholds, one for the within-cluster distance of $\bu_i^\top\bx$ and
$\bu_{i'}^\top\bx$, the other for the between-cluster distance.
} 
We will henceforth call this assumption the 
$\gamma$-gap assumption. We note that such gap assumptions are standard in this literature \cite{glz14,lkg16}.
For user vectors $\bu_1, \ldots, \bu_n \in \R^d$ corresponding to the $n$ users (note that these are unknown to the algorithm), context $\bx \in \R^d$, and user index $i \in \scU$, we denote by $N_i(\bx)$ the {\em true neighborhood} of $i$ w.r.t. $\bx$, i.e., $N_i(\bx) = \{j \in \scU\,:\, \bu_j^\top\bx = \bu_i^\top\bx\}$. Hence, $N_i(\bx)$ is simply the cluster (over $\scU$) that $i$ belongs to w.r.t. $\bx$. Notice that $i \in N_i(\bx)$ for any $i$ and any $\bx$. We will henceforth assume that all instance vectors $\bx$ satisfy $||\bx|| \leq 1$.

As is standard in linear bandit settings~(e.g.,
\cite{Aue02,chu2011contextual,abbasi2011improved,ko11,cg11,yhg12,dkc13,cgz13,ag13,glz14,lkg16,ksl16}, 
and references therein), the unknown user vector $\bu_i$ determines the (average) behavior of user $i$.
More precisely, upon receiving context vector $\bx$, user $i$ ``reacts" by delivering a payoff value\ 
\(
y_i(\bx) = \bu_{i}^\top\bx + \epsilon_{i}(\bx)~,
\)\ 
where $\epsilon_{i}(\bx)$ is a conditionally zero-mean sub-Gaussian error variable with (conditional) variance parameter $\sigma^2(\bx) \leq \sigma^2$ for all $\bx$.\footnote
{
Recall that a zero-mean random variable $X$ is sub-Gaussian with variance parameter $\sigma^2$ if $\E[\exp(sX)] \leq \exp(s^2\,\sigma^2/2)$ for all $s \in \R$. Any variable $X$ with $\E[X] = 0$ and $|X| \leq b$ is sub-Gaussian with variance parameter upper bounded by $b^2$.
}
Hence, conditioned on the past, the quantity $\bu_{i}^\top\bx$ is indeed the expected payoff observed 
at user $i$ for context vector $\bx$.
In fact, for the sake of concreteness, we will assume throughout that for all $i \in \scU$ and $\bx \in \R^d$ we have
$y_i(\bx) \in [-1,+1]$.

As is standard in online learning settings, learning is broken up into a discrete sequence of time steps (or {\em rounds}): 
At each time $t=1,2,\dots$, the learner receives a user index $i_t \in \scU$, representing the user to serve content to. Notice that the user to serve may change from round to round, but the same user may recur several times. Together with $i_t$, the learner receives a set 
of context vectors $C_{t} = \{\bx_{t,1}, \bx_{t,2},\ldots, \bx_{t,c_t}\} \subseteq \R^d$,
such that $||\bx_{t,k}|| \leq 1$ for all $t$ and $k = 1, \ldots, c_t$,
encoding the content which is currently available for recommendation to user $i_t$.
The learner is compelled to pick some ${\bar \bx_t} =  \bx_{t,k_t} \in C_{t}$ 
to recommend to $i_t$, and then observes $i_t$'s feedback in the form of 
payoff $y_t \in [-1,+1]$ whose (conditional) expectation is $\bu_{i_t}^\top{\bar \bx_t}$.
The sequence of pairings $\{i_t,C_t\}_{t=1}^T = \{(i_1,C_1), (i_2,C_2), \ldots, (i_T,C_T)\}$ 
will be generated by an exogenous process and, in a sense, represents the "data at hand". As we shall see  in Section \ref{s:analysis}, the performance of our algorithm will depend on the properties of these data.

The practical goal of the learner is to maximize its total payoff $\sum_{t=1}^T y_t$ over $T$ time
steps. 
From a theoretical standpoint,
we are instead interested in bounding the cumulative {\em regret} achieved by our algorithms. 
More precisely, let the regret $r_t$ of the learner at time $t$ be the extent to which the average 
payoff of the best choice in hindsight at user $i_t$ exceeds the average payoff of the algorithm's choice, 
i.e.,
\[
r_t = \Bigl(\max_{\bx \in C_{t} }\, \bu_{i_t}^\top\bx \Bigl) - \bu_{i_t}^\top{\bar \bx_{t}}~.
\]
We are aimed at bounding with high probability (over the noise variables 
$\epsilon_{i_t}({\bar \bx_{t}})$, and any other
possible source of randomness)
the cumulative regret
\(
\sum_{t=1}^T r_t~.
\)
As a special case of the above model, when the set of items do not possess informative features, we can always resort to the non-contextual bandit setting (e.g., \cite{acf02,ams09}). To implement this approach, we simply take the set of all items (which must be finite for this technique to work), and apply a {\em one-hot} encoding by assigning to the $i$-th item, the $i$-th canonical basis vector $\be_i$, with one at the $i$-th position and zero everywhere else as the context vector. It is easy to see that the expected payoff given by user $i$ on item $j$ will simply be the $j$-th component of vector $\bu_i$. 

Our aim would be to obtain a regret bound that gracefully improves as the context-dependent clustering structure over the users becomes stronger. More specifically, values taken by the number of clusters $m(\bx)$ would be of particular interest since we expect to reap the strongest collaborative effects when $m(\bx)$ is small whereas not much can be done by way of collaborative analysis if $m(\bx) \approx n$. Consequently, a desirable regret bound would be one that diminishes with $m(\bx)$. Yet, recall that $m(\bx)$ is a function of the context vector $\bx$, which means that we expect our regret bound to also depend on the properties of the actual data $\{i_t,C_t\}_{t=1}^T$. We will see in Section \ref{s:analysis} that, under suitable stochastic assumptions on the way $\{i_t,C_t\}_{t=1}^T$ is generated, our regret analysis essentially replaces the dependence on the total number of users $n$ by the (possibly much) smaller quantity $\E[m(\bx)]$, the expected number of clusters over users, the expectation being over the draw of context vectors $\bx$.

\section{The Context-Aware Bandit Algorithm}
\begin{algorithm}[t]
\small{
\begin{algorithmic}[1]
\STATE \textbf{Input}: Separation parameter $\gamma$, exploration parameter $\alpha(t)$
\STATE \textbf{Init}: $\bb_{i,0} = \bzero \in \R^d$ and $M_{i,0} = I \in \R^{d\times d}$,\ \ $i = 1, \ldots n$
\FOR{$t =1,2,\dots,T$}
\STATE Set $\bw_{i,t-1} = M_{i,t-1}^{-1}\bb_{i,t-1}$, \quad for all $i = 1, \ldots, n$
\STATE Use $\CB_{i,t}(\bx) = \alpha(t)\,\sqrt{\bx^\top M_{i,t-1}^{-1} \bx}$, for all $\bx, i = 1,\ldots, n$
\STATE Receive user $i_t \in \scU$, and context vectors $C_{t} = \{\bx_{t,1},\ldots,\bx_{t,c_t}\}$ for items to be recommended\\
\vskip1ex
\texttt{// Compute neighborhoods and aggregates}
\FOR{$k = 1, \ldots, c_t$}
\STATE Compute neighborhood $\hN_{k} := \hN_{i_t,t}(\bx_{t,k})$ for this item\vskip-4ex
\begin{align*}
\hN_{k} = \Bigl\{j \in \scU\,:\, &|\bw_{i_t,t-1}^\top\bx_{t,k} - \bw_{j,t-1}^\top\bx_{t,k}| \\
&\leq \CB_{i_t,t-1}(\bx_{t,k}) + \CB_{j,t-1}(\bx_{t,k}) \Bigr\}\,.
\end{align*}
\STATE \vskip-2ex Set $\bw_{\hN_{k},t-1} = \frac{1}{|\hN_{k}|}\,\sum_{j \in \hN_{k}} \bw_{j,t-1}$
\STATE Set $\CB_{\hN_k,t-1}(\bx_{t,k}) = \frac{1}{|\hN_{k}|}\,\sum_{j \in \hN_{k}} \CB_{j,t-1}(\bx_{t,k})$
\ENDFOR
\STATE \vskip1ex Recommend item ${\bar \bx_{t}} = \bx_{t,k_t} \in C_{t}$ such that\vskip-4ex
\[
k_t = \argmax_{k = 1, \ldots, c_t} \left({\bw_{\hN_{k},t-1}}^\top\bx_{t,k}  + \CB_{\hN_k,t-1}(\bx_{t,k})\right)~.
\]
\STATE \vskip-2ex Observe payoff $y_t \in [-1,1]$~.\\
\vskip1ex
\texttt{// Update user weight vectors}
\IF{$\CB_{i_t,t-1}({\bar \bx_t}) \geq \gamma/4$}
\STATE Set $M_{i_t,t} = M_{i_t,t-1} + {\bar \bx_{t}}{\bar \bx_{t}}^\top$,
\STATE Set $\bb_{i_t,t} = \bb_{i_t,t-1} + y_t {\bar \bx_t}$,
\STATE Set $M_{j,t} =  M_{j,t-1},\ \bb_{j,t} = \bb_{j,t-1}$ for all $j \neq i_t$~.
\ELSE
\FORALL{$j \in \hN_{k_t}$ such that $\CB_{j,t-1}({\bar \bx_t}) < \gamma/4$}
\STATE $M_{j,t} = M_{j,t-1} + {\bar \bx_{t}}{\bar \bx_{t}}^\top$,
\STATE $\bb_{j,t} = \bb_{j,t-1} + y_t {\bar \bx_t}$~.
\ENDFOR
\STATE Set $M_{j,t} =  M_{j,t-1},\ \bb_{j,t} = \bb_{j,t-1}$ for all $j \notin \hN_{k_t}$ and for $j \in \hN_{k_t}$ such that $\CB_{j,t-1}({\bar \bx_t}) \geq \gamma/4$~.
\ENDIF
\ENDFOR
\end{algorithmic}
}
\caption{Context-Aware clustering of Bandits (CAB)}
\label{alg:tcab}
\end{algorithm}
We present Context-Aware (clustering of) Bandits (dubbed as CAB, see Algorithm~\ref{alg:tcab}), an upper-confidence bound-based algorithm for performing recommendations in the context-sensitive bandit clustering model. Similar to previous works \cite{cgz13,glz14,nl14,lkg16,w+16}, CAB maintains a vector estimate $\bw_{i,t}$ to serve as a proxy to the unknown user vector $\bu_i$ at time $t$. CAB also maintains standard correlation matrices $M_{i,t}$. The standard confidence bound function for user $i$ for item $\bx$ at time $t$ is derived as $\CB_{i,t}(\bx) = \alpha(t)\,\sqrt{\bx^\top M_{i,t}^{-1} \bx}$, for a suitable function $\alpha(t) = \scO(\sqrt{d \log t})$.

However, CAB makes sharp departures from previous works both in the way items are recommended, as well as in they way the estimates $\bw_{i,t}$ are updated.

\textbf{Item Recommendation}: At time $t$, we are required to serve user $i_t \in \scU$ by presenting an item out of a set of items $C_{t} = \{\bx_{t,1},\ldots,\bx_{t,c_t}\}$ available at time $t$. To do so, CAB first computes for each item $\bx_{t,k}$ in $C_{t}$, the set of users that are likely to give the item a similar payoff as $i_t$. This set $\hN_{i_t,t}(\bx_{t,k})$ is the \emph{estimated neighborhood} of user $i_t$ with respect to item $\bx_{t,k}$. A user $j$ is included in $\hN_{i_t,t}(\bx_{t,k})$ if the estimated payoff it gives to the item $\bx_{t,k}$ is sufficiently close to that given to the item by user $i_t$ (see step 8).

CAB incorporates collaborative effects by lifting the notions of the user proxy and confidence bounds to a set of users $N \subseteq \scU$. CAB uses a simple, flat averaging lift: $\CB_{N,t}(\bx) = \frac{1}{|N|}\,\sum_{j \in N}\CB_{j,t}(\bx)$ and $\bw_{N,t} =  \frac{1}{|N|} \sum_{j \in N} \bw_{j,t}$. Next, CAB uses (see step 12) aggregated confidence bounds $\CB_{\hN_{i_t,t}(\bx_{t,k})}(\bx_{t,k})$ and aggregated proxy vectors $\bw_{\hN_{i_t,t}(\bx_{t,k}),t-1}$ to select an item ${\bar \bx_t} = \bx_{t,k_t} \in C_{t}$ based on an upper confidence estimation step.


\textbf{Proxy Updates}: Classical approaches update the user proxies $\bw_{i,t}$ by solving a regularized least squares problem involving (feature representations of) items served previously to user $i$ and payoffs received. However, CAB remains fully committed to the collaborative approach (see steps 14-24) by allowing a user $i$ to inherit updates due to an item $\bx$ served to another user $j$ if the two users do indeed agree on their opinion on item $\bx$ with a sufficiently high degree of confidence. After the feedback $y_t$ is received from user $i_t$, the algorithm updates the proxies $\bw_{j,t}$.

If CAB is not too confident regarding the opinion $i_t$ has along the direction ${\bar \bx_t}$, formally $\CB_{i_t,t-1}({\bar \bx_t}) \geq \gamma/4$, then only the proxy at user $i_t$ is updated (see step 15-17). However, if CAB is confident i.e. if $\CB_{i_t,t-1}({\bar \bx_t}) < \gamma/4$ then the proxy updates are performed (see steps 19-23) for all users $j$ in $i_t$'s estimated neighborhood with respect to $\bar\bx_t$ about whose opinions CAB is confident too. Notice that all such users $j$ undergo the same update, which is motivated by the algorithm's belief that $\hN_{i_t,t}({\bar \bx_t}) = N_{i_t}({\bar \bx_t})$, i.e., that the conditional expectation $\bu_{i_t}^\top{\bar \bx_t}$ of $y_t$ given ${\bar \bx_t}$ is actually also equal to $\bu_{j}^\top{\bar \bx_t}$ for all users $j \in \hN_{i_t,t}({\bar \bx_t})$ such that $\CB_{j,t-1}({\bar \bx_t}) < \gamma/4$,

 

It is worth noting that CAB is extremely flexible in handling a fluid set of users $\scU$. Due to its context-sensitive user aggregation step, which is repeated at every round, CAB allows users to be added or dropped on the fly, in a seamless manner. This is in strike contrast to past approaches to bandit aggregation, such as GobLin \cite{cgz13}, CLUB \cite{glz14}, and COFIBA \cite{lkg16}, where more involved feedback sharing mechanisms across the users are implemented which are based either on static network Laplacians or on time-evolving connected components of graphs over a given set of users.

\section{Regret Analysis}
\label{s:analysis}
%
Our regret analysis depends on a specific measure of hardness of the data at hand:
%
for an observed sequence of users $\{i_t\}_{t = 1}^T = \{i_1, \ldots, i_T\}$ and corresponding sequence of item sets $\{C_t\}_{t = 1}^T = \{C_1, \ldots, C_T\}$, where $C_t = \{\bx_{t,1}, \ldots, \bx_{t,c_t}\}$, the {\em hardness} $\HD(\{i_t,C_t\}_{t = 1}^T,\eta)$ of the pairing $\{i_t,C_t\}_{t = 1}^T$ at level $\eta > 0$ is defined as
%
{\small
\begin{align*}
&{\normalsize \HD(\{i_t,C_t\}_{t = 1}^T,\eta)}\\ 
&= \max\Bigl\{t = 1, \ldots, T\,:\, \exists j \in \scU,\,\,\exists k_1,k_2,\ldots, k_t,\,:\\ 
&\ \ \ \ \ I + \sum_{s\leq t\,:\,i_s=j}^t \bx_{s,k_s}\bx_{s,k_s}^\top {\mbox{ has smallest eigenvalue $\leq\eta$}}  \Bigl\}\,.
\end{align*}
}
In words, $\HD(\{i_t,C_t\}_{t = 1}^T,\eta)$ roughly measures the number of rounds we need to wait in the worst case over all possible users $j$ and all possible ways of building matrices $M_{j,t}$ through rank-one adjustments based on the data found in $\{i_t,C_t\}_{t = 1}^T$ until all correlation matrices $M_{j,t}$ have eigenvalues lower bounded by $\eta$.
%
%
Based on the above hardness definition, the following result summarizes our main efforts in this section. The full proof is given in the appendix, along with few ancillary results.
%
\begin{theorem}\label{t:combo}
Let CAB (Algorithm~\ref{alg:tcab}) be run on $\{i_t,C_t\}_{t = 1}^T$, with $c_t \leq c$ for all $t$. Also, let the condition\ \
\(
|\bu_j^\top\bx - \bw_{j,t}^\top\bx| \leq \CB_{j,t}(\bx)
\)
hold for all $j \in \scU$ and $\bx \in \R^d$, along with the $\gamma$-gap assumption. Then the cumulative regret $\sum_{t=1}^T r_t$ of the algorithm can be deterministically upper bounded as follows:
%
{\small
\begin{align*}
\sum_{t=1}^T r_t 
&\leq 9\alpha(T)\Biggl( c\,n\,\HD\Bigl(\{i_t,C_t\}_{t = 1}^T,\frac{16\,\alpha^2(T)}{\gamma^2}\Bigl)\\ 
&\qquad\qquad\qquad\qquad + \sqrt{d\,\log T\,\sum_{t=1}^T \frac{n}{|N_{i_t}({\bar \bx_t})|} }  \Biggl)
\end{align*}
}
\end{theorem}
where we set $\alpha(T) = \scO(\sqrt{\log T})$. Some comments are in order. Theorem \ref{t:combo} delivers a {\em deterministic} regret bound on the cumulative regret, and is composed of two terms. The first term is a measure of hardness of the data sequence $\{i_t,C_t\}_{t = 1}^T$ at hand whereas the second term is the usual $\sqrt{T}$-style term in linear bandit regret analyses \cite{Aue02,chu2011contextual,abbasi2011improved}. However, note that the dependence of the second term on the total number $n$ of users to be served gets replaced by a much smaller quantity $\frac{n}{|N_{i_t}({\bar \bx_t})|}$ that depends on the actual size of context-dependent clusters of the served users.

We will shortly see that if the pairings $\{i_t,C_t\}_{t = 1}^T$ are generated in a favorable manner, such as sampling vectors $\bx_{t,k}$ i.i.d. according to an unknown distribution over the instance space (see Lemma \ref{l:anc1} below), the hardness measure can be upper bounded with high probability by a term of the form $\frac{\log T}{\gamma^2}$. Similarly, for the second term, in the simple case when $N_{i_t}({\bar \bx_t}) = B$ for all $t$, the second term has the form $\sqrt{\frac{n}{B}\,T}$, up to log factors. Notice that $\sqrt{T}$ is roughly the regret effort for learning a single bandit, and $\sqrt{\frac{n}{B}\,T}$ is the effort for learning $\frac{n}{B}$-many (unrelated) clusters of bandits when the clustering is known. Thus, in this example, it is the ratio $\frac{n}{B}$ that quantifies the hardness of the problem, insofar clustering is concerned. Again, under favorable circumstances (see Lemma \ref{l:anc2} below), we can relate the quantity $\sum_{t=1}^T \frac{n}{|N_{i_t}({\bar \bx_t})|}$ to the {\em expected} number of context-dependent clusters of users, the expectation being w.r.t. the random draw of context vectors.

On the other hand, making no assumptions whatsoever on the way $\{i_t,C_t\}_{t = 1}^T$ is generated makes it hard to exploit the cluster structure. For instance, if $\{i_t,C_t\}_{t = 1}^T$ is generated by an adaptive adversary, this might cause $\HD\left(\{i_t,C_t\}_{t = 1}^T,\eta\right)$ to become linear in $T$ for any constant $\eta > 1$, thereby making the bound in Theorem \ref{t:combo} vacuous. However, a naive algorithm that disregards the cluster structure, making no attempts to incorporate collaborative effects, and running $n$-many independent {\sc LinUCB}-like algorithms \cite{Aue02,abbasi2011improved,chu2011contextual}, easily yields a $\sqrt{n\,T}$ regret bound\footnote
{
To see this, simply observe that each of the $n$ {\sc LinUCB}-like algorithms has a regret bound of the form $\sqrt{T_i}$, where $T_i$ is the number of rounds where $i_t = i$. Then $\sum_{t=1}^T r_t \leq \sum_{i=1}^n \sqrt{T_i} \leq \sqrt{n\,T}$, with equality if $T_i = T/n$ for all $i$.
}.

A sufficient condition for controlling the hardness term in Theorem \ref{t:combo} is provided by the following lemma.
%
\begin{lemma}\label{l:anc1}
For each round $t$, let the context vectors $C_t = \{\bx_{t,1}, \ldots, \bx_{t,c_t}\}$ be generated i.i.d. (conditioned on $i_t$, $c_t$, past data $\{i_s,C_s\}_{s=1}^{t-1}$ and rewards $y_1, \ldots, y_{t-1}$) from a sub-Gaussian random vector $X \in \R^d$ with (conditional) variance parameter $\nu^2$, such that $||X|| \leq 1$, and $\E[XX^\top]$ is full rank with smallest eigenvalue $\lambda > 0$. Let also $c_t \leq c$ for all $t$, and $\nu^2 \leq \frac{\lambda^2}{8\ln(4c)}$. Finally, let the sequence $\{i_t\}_{t=1}^T$ be generated uniformly at random,\footnote
{
Any distribution over $\scU$ that assigns a strictly positive probability $p_j$ to all $j \in \scU$ would suffice by replacing $n$ with the inverse of the smallest user probability $p_j$.
} 
independent of all other variables.
Then with probability at least $1-\delta$,
{\small
\[
\HD\Bigl(\{i_t,C_t\}_{t = 1}^T,\eta \Bigl) = \scO\left(\frac{n\,\eta}{\lambda^2}\,\log\left(\frac{Tnd}{\delta}\right)\right)~.
\]
}
\end{lemma}
%
The following lemma 
handles the second term in the bound of Theorem \ref{t:combo}.
\begin{lemma}\label{l:anc2}
For each round $t$, let the context vectors $C_t = \{\bx_{t,1}, \ldots, \bx_{t,c_t}\}$ be generated i.i.d. (conditioned on $i_t$, $c_t$, past data $\{i_s,C_s\}_{s=1}^{t-1}$ and rewards $y_1, \ldots, y_{t-1}$) from a random vector $X \in \R^d$ with $||X|| \leq 1$.
Let also $c_t \leq c$ for all $t$. Then, with probability at least $1-\delta$,
\begin{align*}
\sum_{t=1}^T \frac{1}{|N_{i_t}({\bar \bx_t})|} 
\leq 
\frac{2Tc\,\E[m(X)]}{n} + 12\,\log\left(\frac{\log T}{\delta}\right)~.
\end{align*}
\end{lemma}
%
\begin{remark}\label{r:variance}
The linear dependence on $c$ on the right-hand side can be turned to logarithmic, e.g., at the cost of an extra sub-Gaussian assumption on variables $\frac{1}{|N_{i}(\bx)|}$, $i \in \scU$.
\end{remark}
%
Finally, we recall the following upper confidence bound, from \cite{abbasi2011improved}.
%
\begin{lemma}\label{l:anc3}
Let $\CB_{j,t}(\bx) = \alpha(t)\,\sqrt{\bx^\top M_{j,t}^{-1} \bx}$, with
$\alpha(t) = \scO\left(\sqrt{d\,\log\frac{Tn}{\delta}}\right)$.\footnote
{
The big-oh notation here hides the dependence on the variance $\sigma^2$ of the payoff values.
}  
Then, under the payoff noise model defined in Section \ref{s:prel},\ \ 
\(
|\bu_j^\top\bx - \bw_{j,t}^\top\bx| \leq \CB_{j,t}(\bx)
\)
holds uniformly for all $j \in \scU$, $\bx \in \R^d$, and $t = 1, 2, \ldots $.
\end{lemma}
A straightforward combination of Theorem \ref{t:combo} with Lemmata~\ref{l:anc1},~\ref{l:anc2}, and \ref{l:anc3} yields the following result.
\begin{corollary}\label{c:1}
Let $\CB_{j,t}(\bx)$ be defined with $\alpha(t)$ as in Lemma \ref{l:anc3}, and let the $\gamma$-gap assumption hold. Assume context vectors are generated as in Lemma \ref{l:anc1} 
in such a way that the sub-Gaussian assumption therein
holds
with $c_t \leq c$. Finally, let the sequence $\{i_t\}_{t=1}^T$ be generated as described in Lemma \ref{l:anc1}. Then, with probability at least $1-\delta$, the regret of CAB (Algorithm \ref{alg:tcab}) satisfies
\[
\sum_{t=1}^T r_t 
= 
R+{\tilde \scO}\left(d\sqrt{\,Tc\left(\E[m(X)]\right)} \right)~,
\]
where the ${\tilde \scO}$-notation hides logarithmic factors in $\frac{TNd}{\delta}$, and $R$ is of the form\footnote
{
In fact, no special efforts have been devoted here to finding sharper upper bounds on $R$.
}
\[
R = \frac{c\,n^2\,d\sqrt d}{\lambda^2\,\gamma^2}\,\log^{2.5}\left(\frac{Tnd}{\delta}\right).
\]
\end{corollary}
%
%

\noindent{\bf Sparse user models.} 
We conclude with a pointer to an additional result we have for sparse linear models contained in the supplemental (Section \ref{sec:sparse} therein), which is in line with past analyses on sparse linear bandits for a single user \cite{Abbasi-YadkoriPS2012,CarpentierM2012,Carpentier2015}: If $\bu_1, \ldots, \bu_n$ are $s$-sparse, in the sense that for all $i\in \scU$ it holds that $\|\bu_i\|_0 \leq s$, for $s \ll d$, then replacing the least-squares solution in Step 4 of Figure \ref{alg:tcab} with the solution computed by a two-stage fully corrective method \cite{NeedellT08,DaiM09} allows us to obtain an improved regret bound. Specifically, we can replace factor $d\sqrt{d}$ in $R$ above by $s^2\sqrt{s}$, and factor $d$ multiplying the $\sqrt{T}$-term by a factor of the form $\sqrt{s\,d}$.

%

\section{Experiments}\label{s:exp}
%
We tested CAB on production and real-world datasets, and compared them to standard baselines as well as to state-of-the-art bandit and clustering of bandit algorithms. When no features have been used on the items, a one-hot encoding was adopted. We tried to follow as much as possible previous experimental settings, like those described in~\cite{cgz13,glz14,ksl16,lkg16}.

\subsection{Dataset Description}
\textbf{Tuenti.} Tuenti (owned by Telefonica) is a Spanish social network website that serves ads on its site, the data contains ad impressions viewed by users along with a variable that registers a click on an ad. The dataset contains $d =105$ ads, $n = 14,612$ users, and 15M records/timesteps.
We adopted a one hot encoding scheme for the items, hence items are described by the unit-norm vectors $\be_1, \ldots, \be_d$.
Since the available payoffs are those associated with the items served by the system, we performed offline policy evaluation through a standard importance sampling technique:
we discarded on the fly all records where the system's recommendation (the logged policy) did not coincide with the algorithms' recommendations.
The resulting number of retained records was around $T = 1M$, loosely depending on the different algorithms and runs.
Yet, because this technique delivers reliable estimates when the logged policy makes random choices (e.g., \cite{lcls10}), we actually simulated a random logged policy as follows. At each round $t$, we retained the ad served to the current user $i_t$ with payoff value $a_t$ (1 = ``clicked", 0 = ``not clicked"), but also included $14$ extra items (hence $c_t = 15$ for all $t$) drawn uniformly at random in such a way that, for any item $\be_j$,
if $\be_j$ occurs in some set $C_{t}$, this item will be the one served by system only $1/15$ of the times. Notice that this random selection was independent of the available payoff $a_t$.

\textbf{KDD Cup.}
This dataset was released for the KDD Cup 2012 Online Advertising Competition\footnote{\url{http://www.kddcup2012.org/c/kddcup2012-track2}} where the instances were derived from the session logs of the search engine soso.com. A search session included user, query and ad information, and was divided into multiple instances, each being described using the ad impressed at that time at a certain depth and position. Instances were aggregated with the same user ID, ad ID, and query.
We took the chronological order among all the instances, and seeded the algorithm with the first $c_t = 20$ instances (the length of recommendation lists).
Payoffs $a_t$ are again binary. The resulting dataset had $n=10,333$ distinct users, and $d=6,780$ distinct ads. Similar to the Tuenti dataset, we generated random recommendation lists, and a random logged policy.
We employed one-hot encoding as well in this dataset. The number of retained records was around $T=0,1M$.

\textbf{Avazu.}
This dataset was released for the Avazu Click-Through Rate Prediction Challenge on Kaggle\footnote{\url{https://www.kaggle.com/c/avazu-ctr-prediction}}. Here click-through data were ordered chronologically, and non-clicks and clicks were subsampled according to different strategies. As before, we simulated a random logged policy over recommendation lists of size $c_t = 20$ $\forall t$. Payoffs are once again binary. The final dataset had $n = 48,723$ users, $c_t = 20$ for all $t$, $d=5,099$ items, while the number of retained records was around $T=1,1M$. Again,
we took the one-hot encoding for the items.

\textbf{LastFM and Delicious.}
These two datasets\footnote{\url{www.grouplens.org/node/462}} are extracted from the music streaming service Last.fm and the social bookmarking web service Delicious. The LastFM dataset includes $n$ = 1,892 users, and 17,632 items (the artists). Delicious refers to $n$ = 1,861 users, and 69,226 items (URLs). Preprocessing of data followed previous experimental settings where these datasets have been used, e.g., \cite{cgz13,glz14}. Specifically, after a tf-idf representation of the available items, the context vectors $\bx_{t,i}$ have been generated by retaining only the first $d=25$ principal components. Binary payoffs were created as follows. LastFM: If a user listened to an artist at least once the payoff is 1, otherwise it is 0. Delicious: the payoff is 1 if the user bookmarked the URL, and 0 otherwise. We processed the datasets to make them suitable for use with multi-armed bandit algorithms. Recommendation lists $C_{t}$ of size $c_t = 25$ $\forall t$ were generated at random by first selecting index $i_t$ at random over the $n$ users,
and then padding with 24 vectors chosen at random from the available items up to that time step, in such a way that at least one of these 25 items had payoff 1 for the current user $i_t$. This was repeated for $T = 50,000$ times for the two datasets.

Table \ref{t:stats} summarizes the main statistics of our datasets.
%
\begin{table}[t]
\caption{Dataset statistics. Here, $n$ is the number of users, $d$ is the dimension of the item vectors (which corresponds to the number of items for Tuenti, KDD Cup and Avazu), $c_t$ is the size of the recommendation lists, and $T$ is the number of records (or just {\em retained} records, in the case of Tuenti, KDD Cup and Avazu).}
\label{t:stats}
\vskip -0.25in
\begin{center}
\begin{small}
\begin{sc}
\begin{tabular}{l|rrrr}
Dataset   &$n$ &$d$ &$c_t$ &$T$\\
\hline
Tuenti    &14,612  &105   &15 &$\simeq$1,000,000 \\
KDD Cup   &10,333  &6,780 &20 &$\simeq$100,000 \\
Avazu     &48,723  &5,099 &20 &$\simeq$1,100,000 \\
LastFM    &1,892   &25    &25 &50,000  \\
Delicious &1,861   &25    &25 &50,000  \\
\end{tabular}
\end{sc}
\end{small}
\end{center}
\vskip -0.15in
\end{table}

\begin{figure}[ht]
\begin{center}
\includegraphics[width=0.31\textwidth]{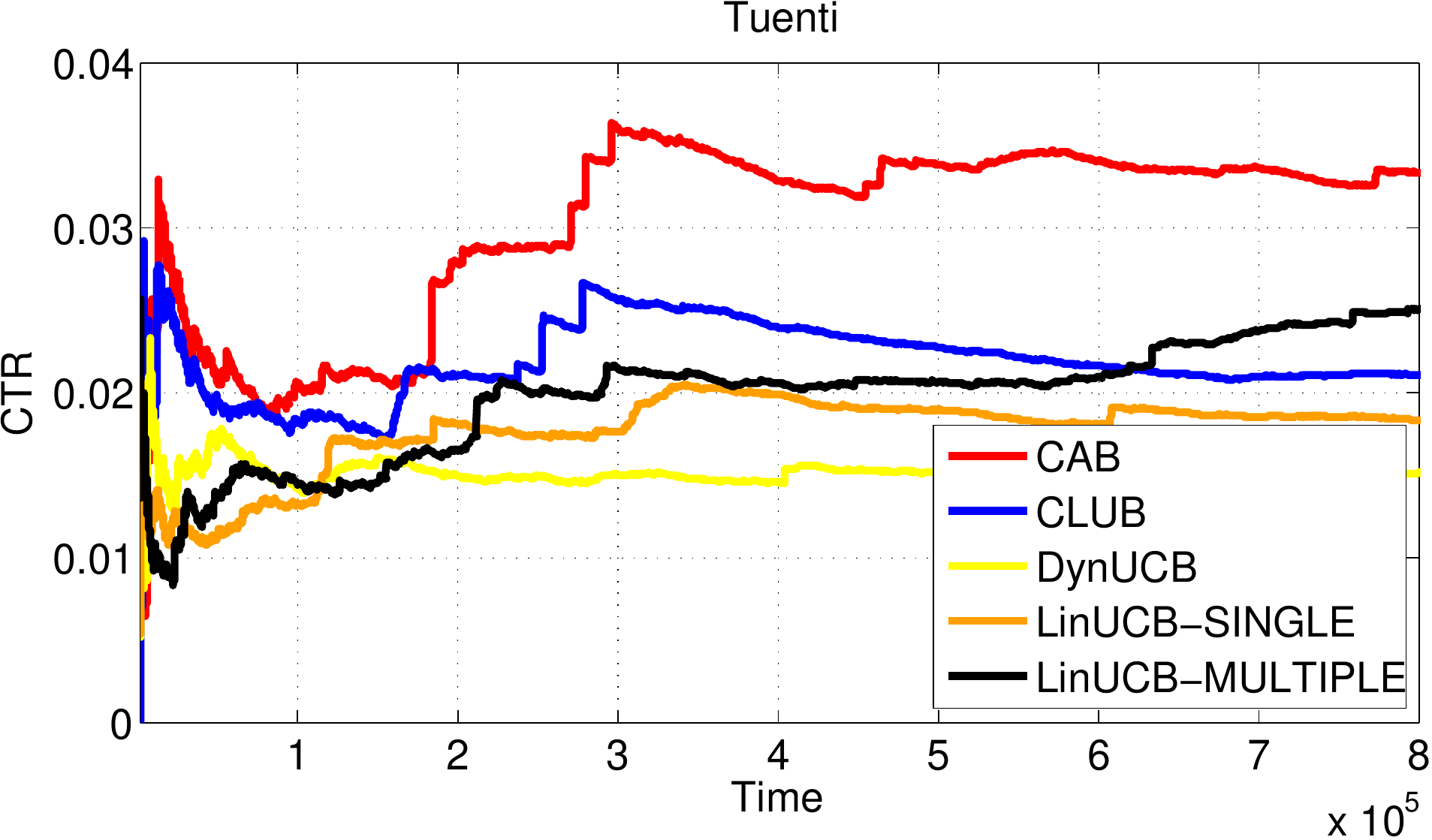}\\
\includegraphics[width=0.31\textwidth]{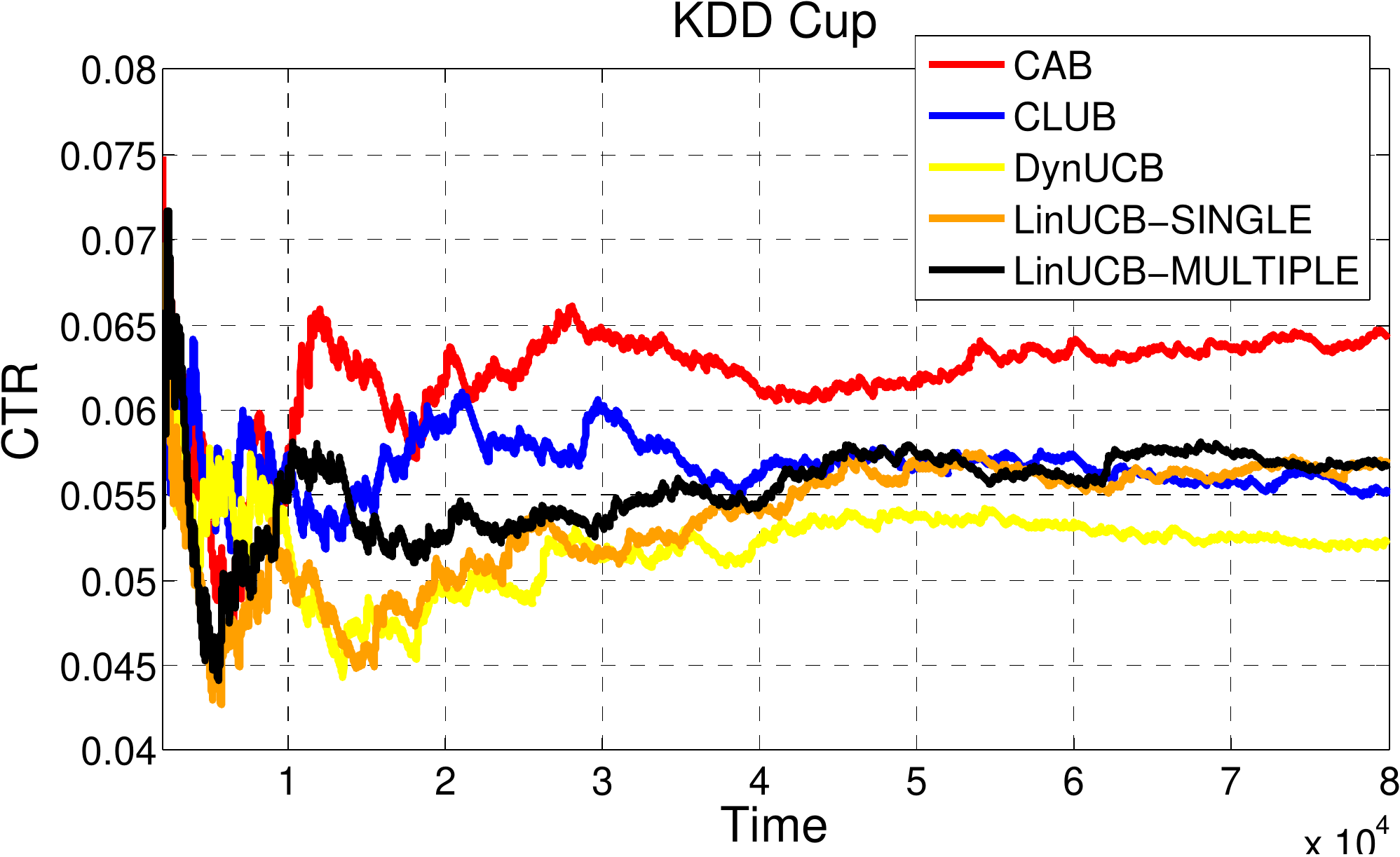}\\
\includegraphics[width=0.31\textwidth]{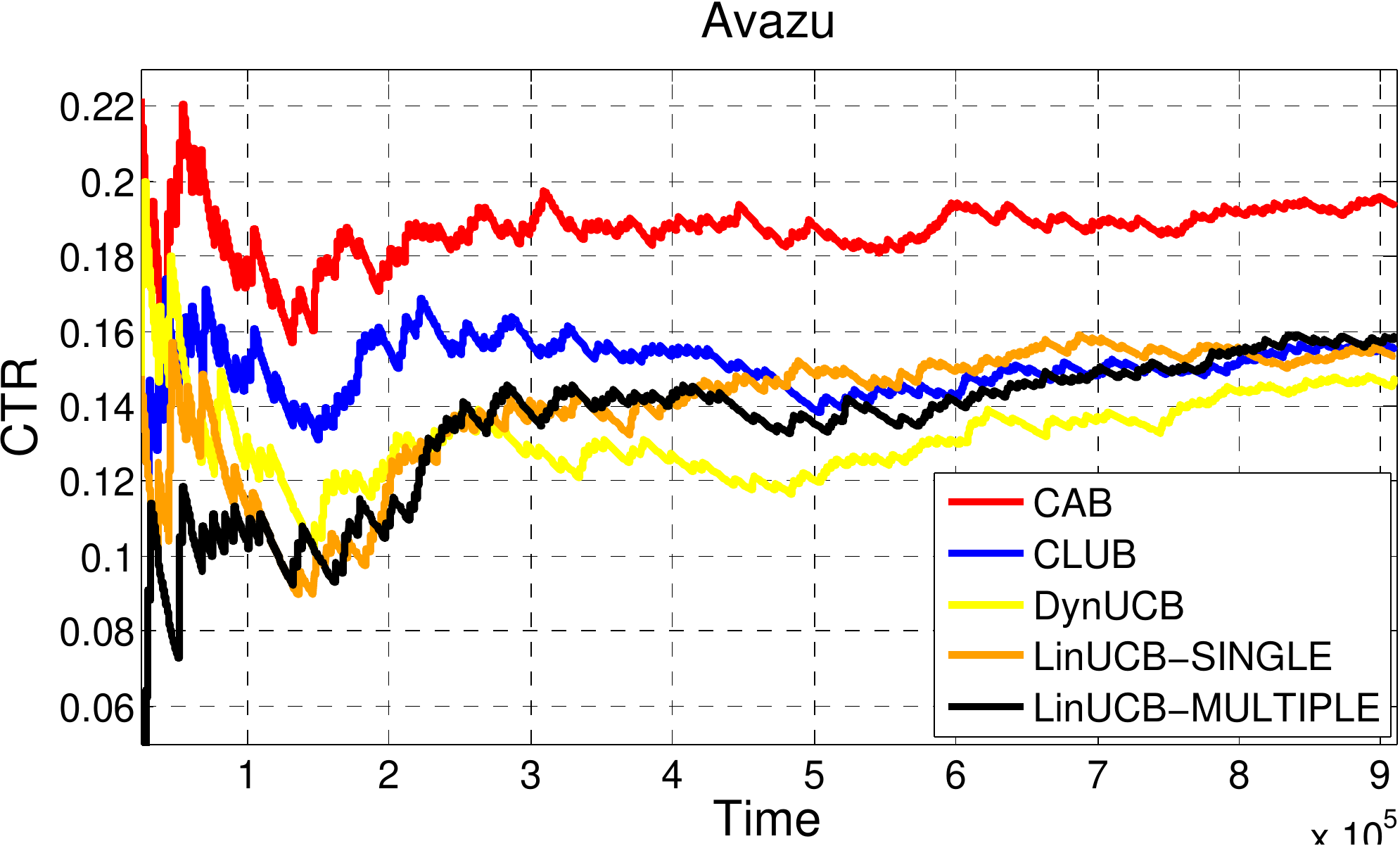}
\caption{Clickthrough Rate vs. retained records ("time") on the three datasets Tuenti, KDD Cup, and Avazu. The higher the curves the better.}
\label{f:results1}
\end{center}
\vskip -0.2in
\end{figure}

\subsection{Algorithms}
%
We used the first $20\%$ of each dataset to tune the algorithms' parameters through a grid search, and report results on the remaining 80\%. All results are averaged over 5 runs.
We compared to a number of state-of-the art bandit and clustering-of-bandit methods:
\begin{itemize}
\item CLUB~\cite{glz14} sequentially refines user clusters based on their confidence ellipsoid balls; We seeded the graph over users by an initial random Erdos-Renyi graphs with sparsity parameter $p = (3\log n)/n$. Because this is a randomized algorithm, each run was repeated five times, and then averaged the results (the observed variance turned out to be small anyway).
%
\item DynUCB~\cite{nl14} uses a traditional $k$-Means algorithm to cluster bandits.
%
\item LinUCB-SINGLE uses a single instance of LinUCB~\cite{chu2011contextual} to serve all users, i.e., all users belong to the same cluster, independent of the items.
%
\item LinUCB-MULTIPLE uses an independent instance of LinUCB per user with no interactions among them, i.e., each user forms a cluster on his/her own, again independent of the items.
%
\item The following variant of CAB (see Algorithm \ref{alg:tcab}): each user $j$ is considered for addition to the estimated neighborhoods $\hN_{k}$ of the currently served user $i_t$ only if $\bw_{j,t-1}$ has been updated at least once in the past.
%
\item Random recommendations, denoted here by RAN, that pick items within $C_{t}$ fully at random.
\end{itemize}
%
%
%
All tested algorithms (excluding RAN) are based on upper-confidence bounds of the form $\CB_{i,t}(\bx) =  \alpha\,\sqrt{\bx^\top N_{i,t}\bx\,\log(1+t)}$. In all cases, we viewed $\alpha$ as a tunable parameter across the values $0, 0.01, 0.02, \ldots, 0.2$. The $\alpha_2$ parameter in CLUB was chosen within $\{0.1, 0,2, \ldots, 0.5\}$. The number of clusters in DynUCB was increased according to an exponential progression, starting from 1, and ending to $n$.
Finally, the $\gamma$ parameter in CAB was simply set to $0.2$. In fact, the value of $\gamma$ did not happen to have a significant influence on the performance of the version of CAB we tested.

\begin{figure}[ht]
\begin{center}
\includegraphics[width=0.30\textwidth]{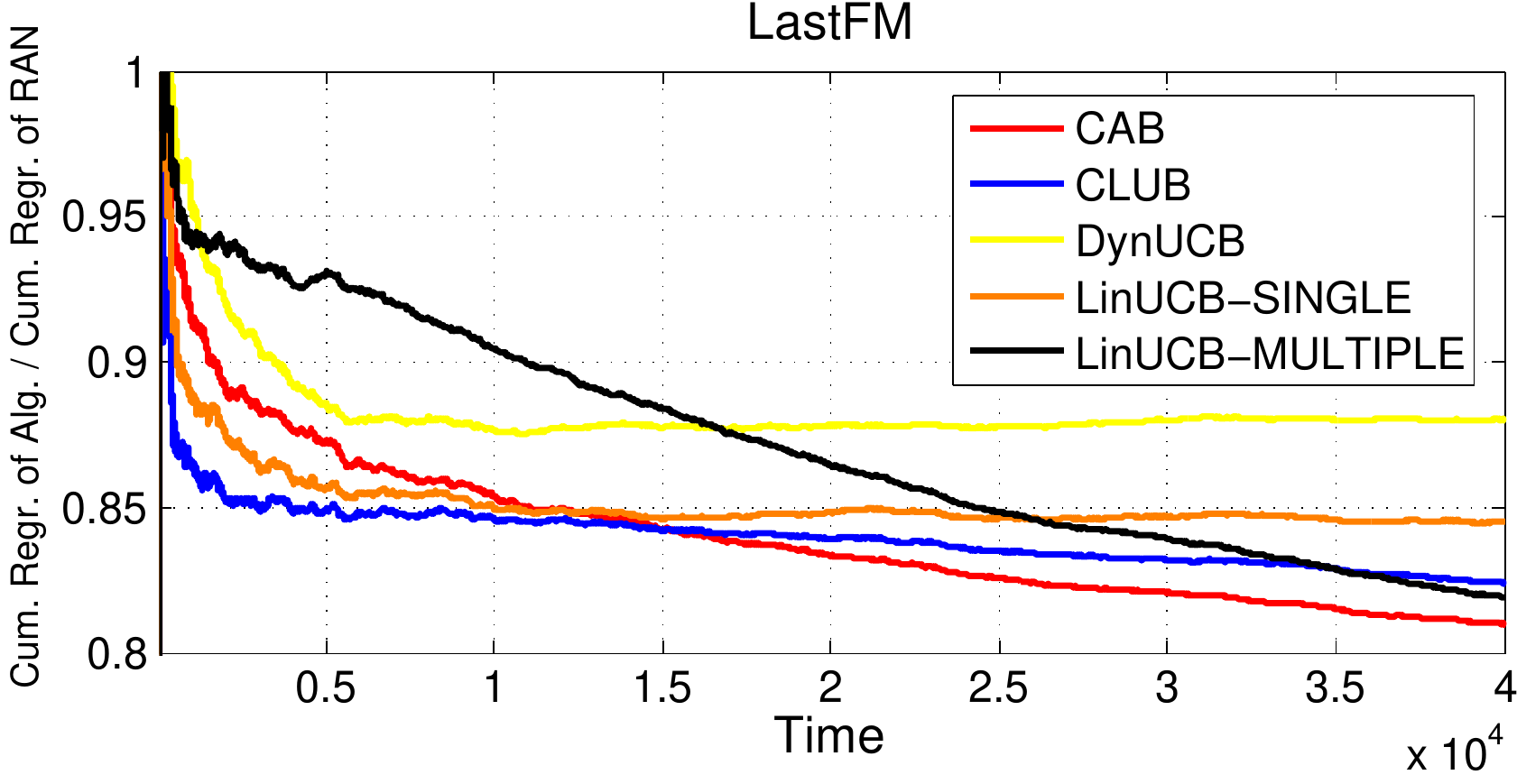}\\
\includegraphics[width=0.30\textwidth]{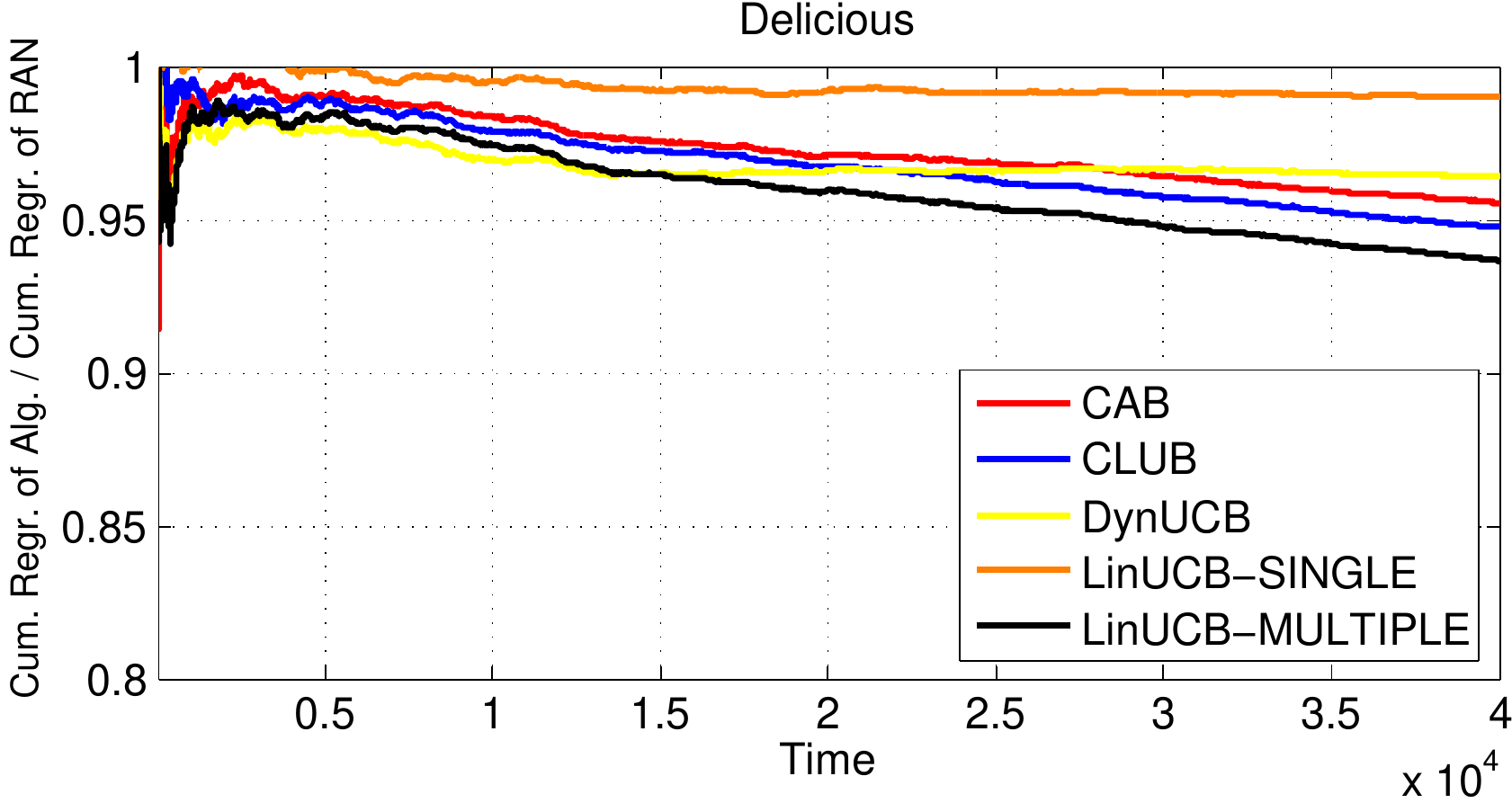}
\caption{Ratio of the cumulative regret of the algorithms to the cumulative regret of RAN against time on the two datasets LastFM and Delicious. The lower the curves the better.}
\label{f:results2}
\end{center}
\vskip -0.1in
\end{figure}

\begin{figure}[ht]
\centering
\hspace*{-0.05in}\includegraphics[width=0.24\textwidth]{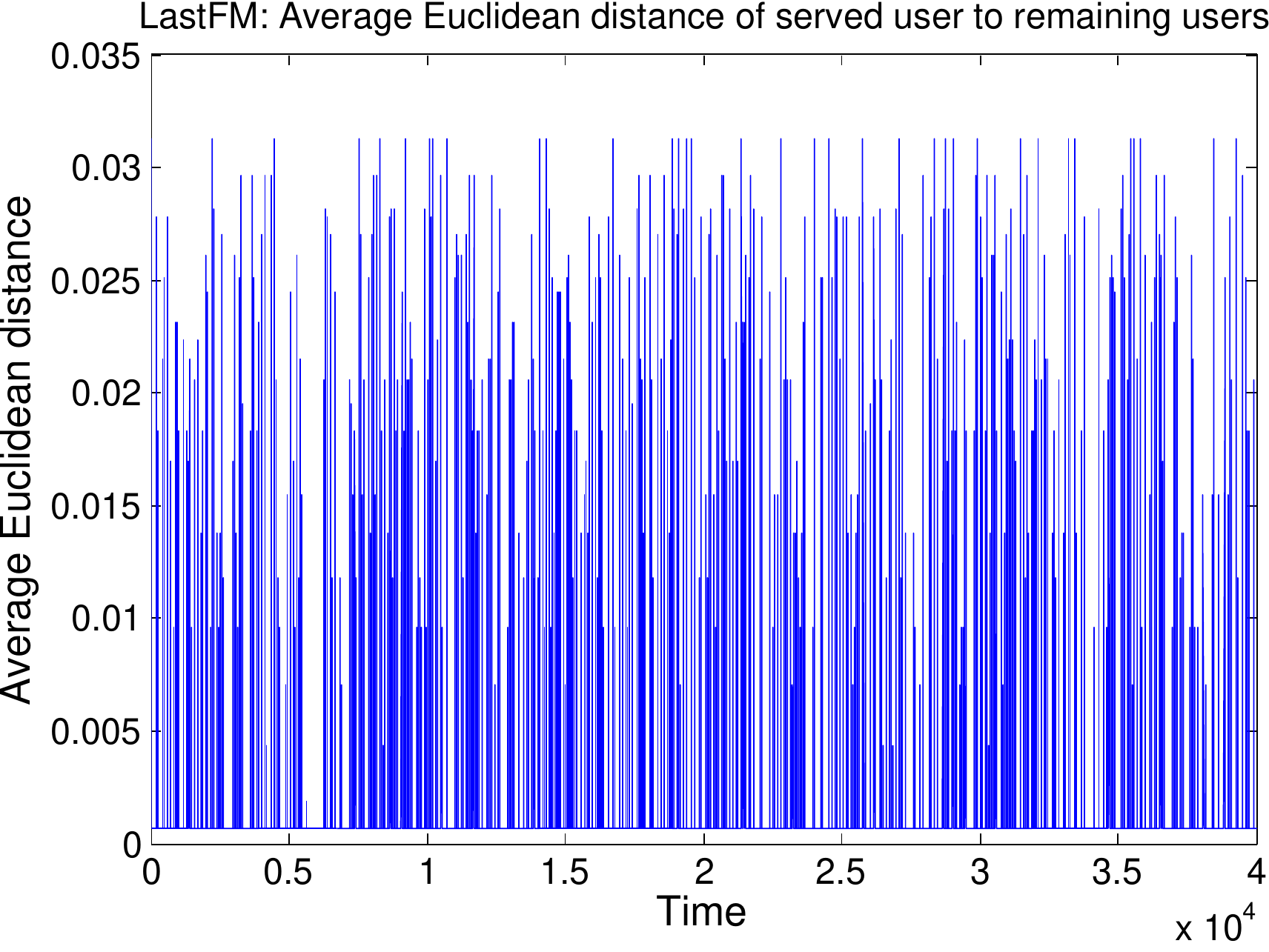}
\hfill
\hspace*{-0.1in}\includegraphics[width=0.24\textwidth]{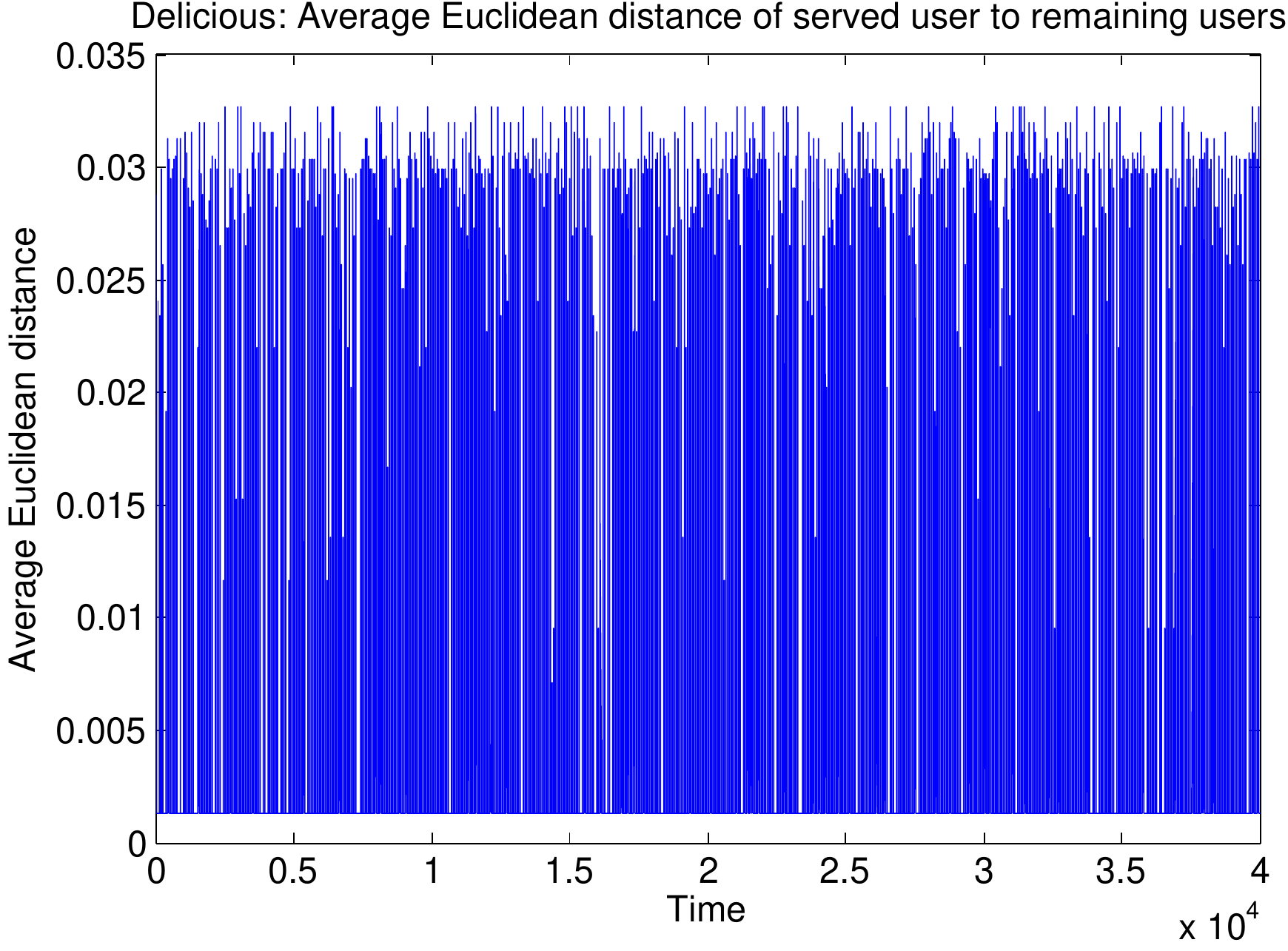}
\caption{Average (estimated) Euclidean distance between the served user $i_t$ and all other users, as a function of $t$ for the two datasets LastFM (left) and Delicious (right). The distance is computed by associating with each user a model vector obtained through a regularized least-squares solution based on all available data for that user (instance vectors and corresponding payoffs).}
\label{f:further}
\vskip -0.2in
\end{figure}

\subsection{Results}
%
The results of our experiments are summarized in Figures \ref{f:results1}, \ref{f:results2}, and \ref{f:further}. All of these results come from the remaining 80\% of the datasets after using 20\% of the data for tuning. For the online advertising datasets Tuenti, KDD Cup, and Avazu (Figure \ref{f:results1}), we measured performance using the Click-Through Rate (CTR), hence the higher the curves the better. For the LastFM and Delicious datasets (Figure \ref{f:results2}), we instead report the ratio of the cumulative regret of the tested algorithm to the cumulative regret of RAN, hence the lower the better.

The experimental setting is in line with past work in the area (e.g., \cite{lcls10,cgz13,glz14,lkg16}), and so are some of the results that we reproduce here. Moreover, by the way data have been prepared, our findings give reliable estimates of the actual CTR performance  (Figure \ref{f:results1}) or actual regret performance (Figure \ref{f:results2}) of the tested algorithms.

In four out of five datasets, CAB was found to offer superior performance, as compared to all baselines. CAB performed particularly well on the Tuenti dataset where it delivers almost double the CTR compared to some of the baselines. CAB's performance advantage was more moderate on the KDD Cup and Avazu datasets. This is expected since exploiting collaborative effects is more important on a dataset like Tuenti, where users are exposed to a few ($\approx$ 100) ads, as compared to the KDD Cup dataset (where ads are also modulated by a user query) and the Avazu dataset, both of which have a much broader ad base ($\approx$ 7000). This provides a strong indication that CAB effectively exploits collaborative effects. In general, on the first three datasets (Tuenti, KDD Cup, and Avazu -- see Figure \ref{f:results1}), CAB was found to offer benefits in the cold-start region (i.e., the initial relatively small fraction of time horizon), but it also continues to maintain a lead throughout.

On the LastFM and Delicious datasets (Figure \ref{f:results2}), the results we report are consistent with \cite{glz14}. On LastFM
all methods are again outperformed by CAB. The overall performance of all bandit methods seems though to be relatively poor; this can be attributed to the way the LastFM dataset was generated. Here users typically have little interaction with the music serving system and a lot of the songs played were generated by a recommender. Hence while there are collaborative effects, they are relatively weak compared to datasets such as Tuenti.

On the other hand, on the Delicious dataset the best performing strategy seems to be LinUCB-MULTIPLE, which deliberately avoids any feedback sharing mechanism among the users.
This dataset reflects user web-browsing patterns, as evinced by their bookmarks. In line with past experimental evidence (e.g., \cite{glz14}), this dataset does not seem to contain any collaborative information, hence we can hardly expect to take advantage of clustering efforts. To shed further light, in Figure \ref{f:further} we plotted the average distance between a linear model for user $i_t$ and the corresponding linear models for all other users $j \neq i_t$, as a function of $t$. For each of the two datasets and each user $i \in \scU$, these linear models have been computed by taking the whole test set and treating each pairing $(\bx_{t,k},y_{t,k})$ with $i_t = i$, and $y_{t,k} = 1$ as a training sample for a (regularized) least-squares estimator for user $i$. The conclusion we can  draw after visually comparing the left and the right plots in Figure \ref{f:further} is that on Delicious these estimated user models tend to be significantly more separated than on LastFM, which easily explains the effectiveness of LinUCB-MULTIPLE. Moreover, on Delicious studies have shown that tags which are used as item features are generally chosen by users to reflect their interests and for personal use, hence we can expect these features to diverge even for similar websites. On the other hand, in LastFM tags are typically reflecting the genre of the song.

\section{Conclusions and Ongoing Research}\label{s:conc}
%
In this paper we proposed a novel contextual bandit algorithm for personalized recommendation systems. Our algorithm is able to effectively incorporate collaborative effects by implementing a simple context-dependent feedback sharing mechanism. Our approach greatly relaxes the restrictions and requirements imposed by earlier works (e.g., \cite{cgz13,glz14,nl14,lkg16,w+16}), and offers a much higher flexibility in handling practical situations, like the on-the-fly inclusion or exclusion of users. Under additional assumptions on the way data are generated, we provided a crisp regret analysis depending on the expected number of clusters over the users, a natural context-dependent notion of the (statistical) difficulty of the learning task. These theoretical findings are further strengthened in the sparse model scenario for users, where improved bounds are shown. We carried out an extensive experimental comparison on a number of production and real-world datasets, with very encouraging results, as compared to available approaches. 

We have started to test (contextual) Thompson Sampling versions of both CAB and its competitors (results are not reported here since they are too preliminary), but so far we have not observed any significant statistical difference compared to what is in Section \ref{s:exp}. From the theoretical standpoint, it would be nice to complement our upper bound in Corollary \ref{c:1} with a lower bound helping to characterize the regret complexity of our problem. From the experimental standpoint, we are planning to have the sparse bandit version of our algorithm undergo a similar experimental validation as the one presented here.

\newpage
\bibliography{biblio}
\bibliographystyle{icml2017}

\clearpage
\newpage

\onecolumn

\appendix

\allowdisplaybreaks
\section{Proofs}
\label{sa:proofs}
The following lemma is the starting point of our regret analysis. 
In what follows, $\{\cdot\}$ denotes the indicator function of the predicate at argument.
\begin{lemma}\label{l:instregret}
Suppose that for all $i \in \scU$, and all $\bx \in \R^d$
it holds that
\[
|\bw_{i,t}^\top\bx - \bu_i^\top\bx| \leq \CB_{i,t}(\bx)\,.
\]
Then the instantaneous regret $r_t$ the CAB algorithm (Algorithm~\ref{alg:tcab}) incurs at time $t$ can be deterministically upper bounded as
\[
r_t \leq (3\alpha(T)+2) \Bigl( \{\hN_{i_t,t}(\bx_t^*) \neq N_{i_t}(\bx_t^*) \} + \{\hN_{i_t,t}({\bar \bx_t}) \neq N_{i_t}({\bar \bx_t})\} \Bigl) + 2\,\CB_{N_{i_t}({\bar \bx_t}),t-1}({\bar \bx_t})~.
\]
\end{lemma}
{\em Proof.} Let $\bx_t^* = \argmax_{k = 1, \ldots, c_t} \bu_{i_t}^\top\bx_{t,k}$, so that
\[
r_t = \bu_{i_t}^\top\bx_t^* - \bu_{i_t}^\top{\bar \bx_t}\,.
\]
Then, setting for brevity
\begin{eqnarray*} 
N^* &=& N_{i_t}(\bx_t^*),\\
\hN^* &=& \hN_{i_t,t}(\bx_t^*),\\ 
N^- &=& N_{i_t}({\bar \bx_t}),\\
\hN^- &=& \hN_{i_t,t}({\bar \bx_t}),
\end{eqnarray*}
we can write
\begin{eqnarray*}
r_t 
&=& \bu_{i_t}^\top\bx_t^* - \bu_{i_t}^\top{\bar \bx_t} \\
&=& \frac{1}{|N^*|}\sum_{j \in N^*}\bu_j^\top\bx_t^* - \bu_{i_t}^\top{\bar \bx_t} \qquad\qquad {\mbox{(since $\bu_j^\top\bx_t^* = \bu_{i_t}^\top\bx_t^*$ for all $j \in N^*$)}}\\
&=& \frac{1}{|N^*|}\sum_{j \in N^*}\Bigl(\bu_j^\top\bx_t^* - \bw_{N^*,t-1}^\top\bx_t^* + \bw_{N^*,t-1}^\top\bx_t^* - \bw_{\hN^*,t-1}^\top\bx_t^* + \bw_{\hN^*,t-1}^\top\bx_t^* + \CB_{\hN^*,t-1}(\bx_t^*) - \CB_{\hN^*,t-1}(\bx_t^*)\\ 
&&\qquad\qquad\qquad - \bu_{i_t}^\top{\bar \bx_t}\Bigl)\,.
\end{eqnarray*}
Using $|\bw_{j,t-1}^\top\bx_t^* - \bu_j^\top\bx_t^*| \leq \CB_{j,t-1}(\bx_t^*)$ for all $j \in N^*$, 
Cauchy-Shwartz inequality, and the definition of ${\bar \bx_t}$, the above can be upper bounded as
\begin{align*}
&\leq\CB_{N^*,t-1}(\bx_t^*) +  ||\bw_{N^*,t-1} - \bw_{\hN^*,t-1}||\cdot \{\hN^* \neq N^* \} + \bw_{\hN^-,t-1}^\top{\bar \bx_t} + \CB_{\hN^-,t-1}({\bar \bx_t}) - \CB_{\hN^*,t-1}(\bx_t^*) - \bu_{i_t}^\top{\bar \bx_t}\\
&= \Bigl(\CB_{N^*,t-1}(\bx_t^*) - \CB_{\hN^*,t-1}(\bx_t^*)\Bigl)\cdot \{\hN^* \neq N^* \} + ||\bw_{N^*,t-1} - \bw_{\hN^*,t-1}||\cdot \{\hN^* \neq N^* \}\\ 
&\qquad\qquad+ \Bigl(\bw_{\hN^-,t-1}^\top{\bar \bx_t} - \bw_{N^-,t-1}^\top{\bar \bx_t}\Bigl)\cdot \{\hN^- \neq N^-\} + \bw_{N^-,t-1}^\top{\bar \bx_t} + \CB_{\hN^-,t-1}({\bar \bx_t}) - \bu_{i_t}^\top{\bar \bx_t}.
\end{align*}
Using again Cauchy-Shwartz inequality, and $\bu_j^\top{\bar \bx_t}  = \bu_{i_t}^\top{\bar \bx_t}$ for all $j \in N^-$, the above can in turn be upper bounded by
\begin{align*}
&\leq \Bigl(\CB_{N^*,t-1}(\bx_t^*) - \CB_{\hN^*,t-1}(\bx_t^*) + ||\bw_{N^*,t-1} - \bw_{\hN^*,t-1}||\Bigl)\cdot \{\hN^* \neq N^* \} + ||\bw_{\hN^-,t-1} - \bw_{N^-,t-1}||\cdot \{\hN^- \neq N^-\}\\
&\qquad\qquad +  \bw_{N^-,t-1}^\top{\bar \bx_t} - \frac{1}{|N^-|}\sum_{j \in N^-} \bu_j^\top{\bar \bx_t} + \CB_{\hN^-,t-1}({\bar \bx_t})\\
&\leq \Bigl(\CB_{N^*,t-1}(\bx_t^*) + ||\bw_{N^*,t-1} - \bw_{\hN^*,t-1}||\Bigl)\cdot \{\hN^* \neq N^* \} + ||\bw_{\hN^-,t-1} - \bw_{N^-,t-1}||\cdot \{\hN^- \neq N^-\} \\
&\qquad\qquad + \CB_{N^-,t-1}({\bar \bx_t}) + \CB_{\hN^-,t-1}({\bar \bx_t})\\
&{\mbox{(using $|\bw_{j,t-1}^\top{\bar \bx_t} - \bu_j^\top{\bar \bx_t}| \leq \CB_{j,t-1}({\bar \bx_t})$ for all $j \in N^-$)}}\\
&= \Bigl(\CB_{N^*,t-1}(\bx_t^*) + ||\bw_{N^*,t-1} - \bw_{\hN^*,t-1}||\Bigl)\cdot \{\hN^* \neq N^* \} + ||\bw_{\hN^-,t-1} - \bw_{N^-,t-1}||\cdot \{\hN^- \neq N^-\} +  2\CB_{N^-,t-1}({\bar \bx_t})\\
&\qquad\qquad + \Bigl(\CB_{\hN^-,t-1}({\bar \bx_t}) - \CB_{N^-,t-1}({\bar \bx_t})\Bigl)\cdot \{\hN^- \neq N^-\}\\
&\leq \Bigl(\CB_{N^*,t-1}(\bx_t^*) + ||\bw_{N^*,t-1} - \bw_{\hN^*,t-1}||\Bigl)\cdot \{\hN^* \neq N^* \} + \Bigl(\CB_{\hN^-,t-1}({\bar \bx_t}) + ||\bw_{\hN^-,t-1} - \bw_{N^-,t-1}||\Bigl)\cdot \{\hN^- \neq N^-\} \\
&\qquad\qquad +  2\CB_{N^-,t-1}({\bar \bx_t})\,.
\end{align*}
We now handle the terms within the round braces. Since $\max_{\bx\,:\, ||\bx||\leq 1} \bx^\top M_{i,t}^{-1} \bx \leq 1$ for all $i$ and $t$ (by construction, $M_{i,t} \succeq I$ as $M_{i,0} = I$), we have that $\CB_{N^*,t-1}(\bx_t^*)$ and $\CB_{\hN^-,t-1}({\bar \bx_t})$ are both upper bounded by $\alpha(T)$. Moreover, using the shorthand $\bu_{N} = \frac{1}{|N|}\,\sum_{j \in N} \bu_j$, for $N \subseteq \scU$, we have
\begin{align*}
||\bw_{N^*,t-1} - \bw_{\hN^*,t-1}||
&\leq
||\bw_{N^*,t-1} - \bu_{N^*}|| + ||\bu_{N^*} - \bu_{\hN^*}|| + ||\bu_{\hN^*} - \bw_{\hN^*,t-1}||\\
&\leq \max_{\bx\,:\, ||\bx||\leq 1} \CB_{N^*,t-1}(\bx) + ||\bu_{N^*}|| + ||\bu_{\hN^*}|| + \max_{\bx\,:\, ||\bx||\leq 1} \CB_{\hN^*,t-1}(\bx)\\
&\leq 2(\alpha(T)+1)~.
\end{align*}
Hence, we conclude that
\begin{align*}
r_t &\leq (3\alpha(T)+2)\left(\{\hN^* \neq N^* \} + \{\hN^- \neq N^-\} \right) +  2\CB_{N^-,t-1}({\bar \bx_t})~,
\end{align*}
as claimed. $\hfill\Box$\\

Under the $\gamma$-gap assumption, we also have the following lemma.
\begin{lemma}\label{l:1}
Let $i_t$ be the user served at time $t$ (see Figure \ref{alg:tcab}). Let
\[
|\bu_j^\top\bx - \bw_{j,t}^\top\bx| \leq \CB_{j,t}(\bx)
\]
hold for all $j \in \scU$,
and $\bx \in \R^d$. Also, for fixed $\bx_t^o$, let $\CB_{j,t-1}(\bx_t^o) < \gamma/4$ holds for all $j \in \scU$. Then
\[
\hN_{i_t,t}(\bx_t^o) = N_{i_t}(\bx_t^o)~.
\]
\end{lemma}
{\em Proof.}
We first claim that, under the assumptions of this lemma, the following two implications hold:
\begin{enumerate}
\item[1.] Given $i,j \in \scU$, if $\bu_i^\top\bx \neq \bu_j^\top\bx$ and $\CB_{i,t}(\bx) + \CB_{j,t}(\bx) < \gamma/2$ then 
$
|\bw_{i,t}^\top\bx - \bw_{j,t}^\top\bx| > \CB_{i,t}(\bx) + \CB_{j,t}(\bx)~.
$
\item[2.] Given $i,j \in \scU$, if $|\bw_{i,t}^\top\bx - \bw_{j,t}^\top\bx| > \CB_{i,t}(\bx) + \CB_{j,t}(\bx)$ then
$
\bu_i^\top\bx \neq \bu_j^\top\bx~.
$
\end{enumerate}
In order to prove Item 1, notice that the $\gamma$-gap assumption entails that $\bu_i^\top\bx \neq \bu_j^\top\bx$ is equivalent to $|\bu_i^\top\bx - \bu_j^\top\bx| \geq \gamma$. Hence we can write
\begin{align*}
\gamma 
&\leq |\bu_{i}^\top\bx - \bu_{j}^\top\bx| \\
&\leq |\bu_{i}^\top\bx - \bw_{i,t}^\top\bx| + |\bw_{i,t}^\top\bx - \bw_{j,t}^\top\bx| + |\bw_{j,t}^\top\bx - \bu_{j}^\top\bx|\\
&\leq \CB_{i,t}(\bx) + |\bw_{i,t}^\top\bx - \bw_{j,t}^\top\bx| + \CB_{j,t}(\bx)\\
&< \gamma/2 + |\bw_{i,t}^\top\bx - \bw_{j,t}^\top\bx|~,
\end{align*}
implying that
\[
|\bw_{i,t}^\top\bx - \bw_{j,t}^\top\bx| > \gamma/2 > \CB_{i,t}(\bx) + \CB_{j,t}(\bx)~.
\]
As for Item 2, we can write
\begin{align*}
\CB_{i,t}(\bx)  + \CB_{j,t}(\bx) &< |\bw_{i,t}^\top\bx - \bw_{j,t}^\top\bx|\\ 
&\leq |\bw_{i,t}^\top\bx - \bu_{i}^\top\bx| + |\bu_{i}^\top\bx - \bu_{j}^\top\bx| + |\bu_{j}^\top\bx - \bw_{j,t}^\top\bx|\\
&\leq \CB_{i,t}(\bx) + |\bu_{i}^\top\bx - \bu_{j}^\top\bx| + \CB_{j,t}(\bx)~,
\end{align*}
implying that $|\bu_{i}^\top\bx - \bu_{j}^\top\bx| >0$. Using the above two claims, we want to show that both
\begin{enumerate}
\item[1a.] $N_{i_t}(\bx_t^o) \subseteq \hN_{i_t,t}(\bx_t^o)$ and 
\item[2a.] $\hN_{i_t,t}(\bx_t^o) \subseteq N_{i_t}(\bx_t^o)$
\end{enumerate}
hold. We choose $i = i_t$ in the above. Then, in order to prove Item 1a, we observe that if $j \in \scU$ is such that $\bu_{i_t}^\top\bx_t^o = \bu_{j}^\top\bx_t^o$ then Item 2 above implies $|\bw_{i_t,t-1}^\top\bx_t^o - \bw_{j,t-1}^\top\bx_t^o| \leq \CB_{i_t,t-1}(\bx_t^o) + \CB_{j,t-1}(\bx_t^o)$, i.e., $j \in \hN_{i_t,t}(\bx_t^o)$. On the other hand, if $j$ is such that $|\bw_{i_t,t-1}^\top\bx_t^o - \bw_{j,t-1}^\top\bx_t^o| \leq \CB_{i_t,t-1}(\bx_t^o) + \CB_{j,t-1}(\bx_t^o)$ then Item 1 above allows us to conclude that either $\bu_{i_t}^\top\bx_t^o = \bu_j^\top\bx_t^o$ or $\CB_{i_t,t-1}(\bx_t^o) + \CB_{j,t-1}(\bx_t^o) \geq \gamma/2$. Yet, because $\CB_{j,t-1}(\bx_t^o) < \gamma/4$ for all $j \in \scU$, the second conclusion is ruled out, thereby implying $j \in N_{i_t}(\bx_t^o)$.  $\hfill\Box$

\begin{remark}\label{r:1}
It is important to observe that, under the hypotheses of Lemma \ref{l:1} (when setting there $\bx_t^o = {\bar \bx_t}$), also the set of $j \in \scU $ whose profile $\bw_{j,t-1}$ gets updated at the end of round $t$ in Figure \ref{alg:tcab} coincides with $N_{i_t}({\bar \bx_t})$.
\end{remark}
By setting $\bx_t^o$ to either $\bx_t^*$ or ${\bar \bx_t}$, we now combine Lemma \ref{l:1} with Lemma \ref{l:instregret} to bound the number of rounds $t$ such that $\hN_{i_t,t}(\bx_t^*) \neq N_{i_t}(\bx_t^*)$ and the number of rounds $t$ such that $\hN_{i_t,t}({\bar \bx_t}) \neq N_{i_t}({\bar \bx_t})$. In turn, these will be immediately related to the hardness $\HD(\{i_t,C_t\}_{t = 1}^T,\eta)$ of the data $\{i_t,C_t\}_{t = 1}^T$ at our disposal.
Moreover, we will use Remark \ref{r:1} to exploit the fact that when the confidence bounds $\CB_{j,t-1}({\bar \bx_t})$ are all small enough along the selected direction ${\bar \bx_t}$, then the number of weight updates performed in round $t$ is exactly equal to the size of the true neighborhood $N_{i_t}({\bar \bx_t})$. 

{\bf Proof of Theorem \ref{t:combo}.} 
Consider the bound in Lemma \ref{l:instregret}. We can write
\begin{align*}
\{\hN_{i_t,t}(\bx_t^*) \neq N_{i_t}(\bx_t^*) \} &\leq \{\exists k = 1, \ldots, c_t\,:\,\hN_{i_t,t}(\bx_{t,k}) \neq N_{i_t}(\bx_{t,k}) \} \\
&\leq 
\sum_{k = 1}^{c_t} \{ \hN_{i_t,t}(\bx_{t,k}) \neq N_{i_t}(\bx_{t,k}) \} \\
&\leq 
\sum_{k = 1}^{c_t} \{\exists j \in \scU\,:\, \CB_{j,t-1}(\bx_{t,k}) \geq \gamma/4 \} \qquad {\mbox{(using Lemma \ref{l:1})}} \\
&\leq 
\sum_{k = 1}^{c_t} \sum_{j \in \scU} \{\CB_{j,t-1}(\bx_{t,k}) \geq \gamma/4 \}\,.
\end{align*}
Clearly, the very same upper bound applies to $\{\hN_{i_t,t}({\bar \bx_t}) \neq N_{i_t}({\bar \bx_t})\}$. Moreover,
\begin{align*}
\sum_{t=1}^T \CB_{N_{i_t}({\bar \bx_t}),t-1}({\bar \bx_t}) &= \sum_{t=1}^T \{\exists j \exists k\,\,:\,\CB_{j,t-1}(\bx_{t,k}) \geq \gamma/4\} \times\CB_{N_{i_t}({\bar \bx_t}),t-1}({\bar \bx_t}) + \sum_{t\,:\,\CB_{j,t-1}(\bx_{t,k}) < \gamma/4\,\forall j \forall k} \CB_{N_{i_t}({\bar \bx_t}),t-1}({\bar \bx_t})\\
&\leq
\sum_{t=1}^T \sum_{k=1}^{c_t} \sum_{j \in \scU}\{\CB_{j,t-1}(\bx_{t,k}) \geq \gamma/4\} \times\max_{\bx\,:\,||\bx||\leq 1}\CB_{N_{i_t}({\bar \bx_t}),t-1}(\bx)\\
&\qquad\qquad + \sum_{t\,:\,\CB_{j,t-1}(\bx_{t,k}) < \gamma/4\,\forall j \forall k} \CB_{N_{i_t}({\bar \bx_t}),t-1}({\bar \bx_t})\\
&\leq
\alpha(T)\,\sum_{t=1}^T \sum_{k=1}^{c_t} \sum_{j \in \scU}\{\CB_{j,t-1}(\bx_{t,k}) \geq \gamma/4\} + \sum_{t\,:\,\CB_{j,t-1}(\bx_{t,k}) < \gamma/4\,\forall j \forall k} \CB_{N_{i_t}({\bar \bx_t}),t-1}({\bar \bx_t})\,.
\end{align*}
Putting together as in Lemma \ref{l:instregret} gives 
\begin{align}\label{e:prebound}
\sum_{t=1}^T r_t 
&\leq 
(8\alpha(T)+4)\,\sum_{t=1}^T \sum_{k=1}^{c_t}\sum_{j \in \scU} \{\CB_{j,t-1}(\bx_{t,k}) \geq \gamma/4 \} + 2\,\sum_{t\,:\,\CB_{j,t-1}(\bx_{t,k}) < \gamma/4\,\forall j\,\forall k } \CB_{N_{i_t}({\bar \bx_t}),t-1}({\bar \bx_t})\,. 
\end{align}
We first focus on the triple sum in (\ref{e:prebound}), which is easily rewritten in terms of $\HD(\{i_t,C_t\}_{t = 1}^T,\eta)$, for a suitable level $\eta$. In fact, if we denote by $\lambda_{\max}(\cdot)$ and $\lambda_{\min}(\cdot)$ the maximal and the minimal eigenvalue of the matrix at argument, we have
\begin{align*}
\CB_{j,t-1}(\bx_{t,k}) 
&= \alpha(t)\,\sqrt{\bx_{t,k}^\top M_{j,t-1}^{-1}\bx_{t,k}}\\
&\leq \alpha(T)\,\sqrt{\lambda_{\max}(M_{j,t-1}^{-1})}\\
&= \frac{\alpha(T)}{\sqrt{\lambda_{\min}(M_{j,t-1})}}\,,
\end{align*}
which is smaller than $\gamma/4$ if $\lambda_{\min}(M_{j,t-1}) > \frac{16\,\alpha^2(T)}{\gamma^2}$. Hence, recalling that $n = |\scU|$ and $c_t \leq c$ for all $t$,
\begin{align*}
\sum_{t=1}^T &\sum_{k=1}^{c_t}\sum_{j \in \scU} \{\CB_{j,t-1}(\bx_{t,k}) \geq \gamma/4 \} \leq
c\,n\,\HD\left(\{i_t,C_t\}_{t = 1}^T,\frac{16\,\alpha^2(T)}{\gamma^2}\right)\,.
\end{align*}
Next, we focus on the last sum in (\ref{e:prebound}). Let $\up(j) \subseteq \{1, \ldots, T\}$ be the set of rounds $t$ such that $\bw_{j,t}$ undergoes an update. Also, let $G = \{t = 1, \ldots, T\,:\,\CB_{j,t-1}({\bar \bx_t}) < \gamma/4\,\forall j \in \scU \}$. Notice that, for all $j \in \scU$ and $t = 1, \ldots, T$,
\begin{equation}\label{e:M}
M_{j,t} = I + \sum_{s \leq t\,:\,s\in \up(j)}{\bar \bx_s}{\bar \bx_s}^\top\,.
\end{equation}
We can write
\begin{align}
\sum_{t\,:\,\CB_{j,t-1}(\bx_{t,k}) < \gamma/4\,\forall j\,\forall k } \CB_{N_{i_t}({\bar \bx_t}),t-1}({\bar \bx_t}) &\leq \sum_{t\in G} \CB_{N_{i_t}({\bar \bx_t}),t-1}({\bar \bx_t}) \nonumber\\
&= \sum_{t\in G} \frac{1}{|N_{i_t}({\bar \bx_t})|}\,\sum_{j\in N_{i_t}({\bar \bx_t})} \CB_{j,t-1}({\bar \bx_t})  \nonumber\\
&= \sum_{j \in \scU}\,\,\sum_{t\in G :j \in N_{i_t}({\bar \bx_t}) } \frac{\CB_{j,t-1}({\bar \bx_t})}{|N_{i_t}({\bar \bx_t})|} \nonumber\\
&\leq \sum_{j \in \scU}\,\left(\sum_{t\in G:j \in N_{i_t}({\bar \bx_t})} \frac{1}{|N_{i_t}({\bar \bx_t})|^2} \right)^{1/2} \times \left(\sum_{t\in G:j \in N_{i_t}({\bar \bx_t})} \CB^2_{j,t-1}({\bar \bx_t})\right)^{1/2} \label{e:twoparts}\\
&{\mbox{(from Cauchy-Shwartz inequality)\,.}} \nonumber
\end{align}
Now, observe that from Remark \ref{r:1}, $\{t\in G:j \in N_{i_t}({\bar \bx_t})\} \subseteq \up(j)$. Hence, for each $j \in \scU$, we have
\begin{align*}
\sum_{t\in G:j \in N_{i_t}({\bar \bx_t})} \CB^2_{j,t-1}({\bar \bx_t}) &\leq \sum_{t\in \up(j)} \CB^2_{j,t-1}({\bar \bx_t})\\
&\leq
\alpha^2(T)\sum_{t \in \up(j)}{\bar \bx_t}^\top M_{j,t-1}^{-1}{\bar \bx_t}\\
&\leq
2\,\alpha^2(T)\log\frac{|M_{j,T}|}{|M_{j,0}|}\\
&{\mbox{(from, e.g., Lemma 24 in \cite{dgs12})}}\\
&\leq
2\,d\,\alpha^2(T)\log(1+ |\up(j)|)\\ 
&\leq  2\,d\,\alpha^2(T)\log(1+T)\,,
\end{align*}
where $|M|$ denotes the determinant of matrix $M$.
Furthermore, again from Cauchy-Shwartz inequality, we can write
\begin{align*}
\sum_{j \in \scU}\,\Biggl(\sum_{t\in G:j \in N_{i_t}({\bar \bx_t})} \frac{1}{|N_{i_t}({\bar \bx_t})|^2} \Biggl)^{1/2} &\leq \left(n\,\sum_{j \in \scU}\,\sum_{t\in G:j \in N_{i_t}({\bar \bx_t})}\frac{1}{|N_{i_t}({\bar \bx_t})|^2} \right)^{1/2}\\
&=
\left(n\,\sum_{t\in G}\sum_{j \in N_{i_t}({\bar \bx_t})} \frac{1}{|N_{i_t}({\bar \bx_t})|^2} \right)^{1/2} \leq
\left(n\,\sum_{t=1}^T \frac{1}{|N_{i_t}({\bar \bx_t})|} \right)^{1/2}\,.
\end{align*}
Piecing together as in (\ref{e:twoparts}), and plugging back into (\ref{e:prebound}) gives the claimed result. $\hfill\Box$

{\em Proof sketch of Lemma \ref{l:anc1}. } 
The proof is similar to that of Lemma 2 in \cite{glz14}, where it is shown (Claim 1 therein) that under the assumptions of this lemma 
\begin{align*}
\E_t\left[\min_{k \in \{1\,\ldots, c_t\} }(\bz^\top\bx_{t,k})^2\,|\,(i_t,c_t) \right] \geq \lambda/4\,. 
\end{align*}
The proof then continues as in Lemma 2 of \cite{glz14} by setting up a Freedman-style matrix tail bound to get, as a consequence of the above, the following high-confidence estimate, holding with probability at least $1-\delta$, uniformly over $j \in \scU$, and $t = 1, 2, \ldots, $:
\begin{align}
\min_{k_1 \in \{1,\ldots,c_1\}, \ldots, k_t \in \{1,\ldots,c_t\}} 
\lambda_{\min}\left(I + \sum_{s \leq t\,:\,i_s = j} \bx_{s,k_s} \bx_{s,k_s}^\top \right) \geq 1 + B_{\lambda,\nu,c}\left(T_{j,t},\frac{\delta}{2nd}\right)~,\label{e:mineigen}
\end{align}
where
\begin{align*}
B_{\lambda,\nu,c}&(T,\delta) = \lambda/4\,T - 8\left(\log(T/\delta) + \sqrt{T\,\log(T/\delta)} \right)\,.
\end{align*}
We continue by lower bounding (\ref{e:mineigen}) with high probability. Observe that, for any fixed $j$ and $t$, variable $T_{j,t}$ is binomial with parameters $t$ and $1/n$. Let us define the auxiliary function
\[
D(x) = 2n\left(x+\frac{5}{3}\log\left(\frac{Tn}{\delta}\right) \right) 
= \scO\left( n x+n\log\left(\frac{Tn}{\delta}\right) \right)~.
\]
A standard application of Bernstein inequality to (Bernoulli) i.i.d. sequences allows us to conclude that, for any fixed value $x$,
\begin{equation}\label{e:bern}
\Pr(\exists j \in \scU,\,\exists t > D(x)\,:\, T_{j,t} \leq x) \leq \delta~.
\end{equation}
Now, in order for (\ref{e:mineigen}) to be lower bounded by $\eta$ for all $j \in \scU$ with probability at least $1-\delta$, it suffices to have
\[
T_{j,t} = \Omega \left(\frac{\eta}{\lambda^2}\,\log\left(\frac{nd}{\delta}\right)\right) := x^*~.
\] 
We set $x = x^*$ into (\ref{e:bern}) to conclude that when
\[
t \geq D(x^*) = \Omega \left(\frac{n\,\eta}{\lambda^2}\,\log\left(\frac{Tnd}{\delta}\right)\right)
\]
then
\[
\HD\Bigl(\{i_t,C_t\}_{t = 1}^T,\eta \Bigl) = \scO\left(\frac{n\,\eta}{\lambda^2}\,\log\left(\frac{Tnd}{\delta}\right)\right)~,
\]
as claimed $\hfill\Box$

{\em Proof of Lemma \ref{l:anc2}. } 
Fix round $t$, let
$\E_t[\cdot]$ denote the conditional expectation $\E\left[\cdot\,|\,\{i_s,C_s\}_{s=1}^{t-1}, y_1, \ldots, y_{t-1} \right]$.
%
We have
\begin{align*}
\E_t\Bigl[\frac{1}{|N_{i_t}({\bar \bx_t})|} \,|\, (i_t,c_t) \Bigl] &\leq \E_t\left[\sum_{k = 1}^{c_t} \frac{1}{|N_{i_t}(\bx_{t,k})|} \,|\, (i_t,c_t) \right]~,
\end{align*}
so that
\begin{align}
\E_t\left[\frac{1}{|N_{i_t}({\bar \bx_t})|} \,|\, c_t \right] 
&\leq
\frac{1}{n}\,\sum_{i=1}^n \E_t\left[\sum_{k = 1}^{c_t} \frac{1}{|N_{i}(\bx_{t,k})|}\,|\, c_t \right]\notag\\
&= \frac{1}{n}\sum_{k=1}^{c_t}\E_t\left[\sum_{i=1}^n \frac{1}{|N_{i}(\bx_{t,k})|}\,|\, c_t \right]  \notag\\
&= \frac{c_t}{n}\E\left[m(X)\right]\notag\\
&\leq \frac{c}{n}\E\left[m(X)\right]\,,\label{e:expectation}
\end{align}
the last equality deriving from the fact that for any given $\bx$, the set of $n$ users is partitioned into $m(\bx)$ clusters corresponding to the neighborhoods $N_i(\bx)$ (so that
$\sum_{i=1}^n \frac{1}{|N_{i}(\bx)|} = m(\bx)$).
Let us now define the variables
\[
D_t = \frac{1}{|N_{i_t}({\bar \bx_t})|} -  \E_t\left[\frac{1}{|N_{i_t}({\bar \bx_t})|}\,|\, c_t \right]~,
\] 
for $t = 1,\ldots, T$. We have that $D_1, \ldots, D_T$ is a martingale difference sequence to which we can apply standard concentration inequalities. In particular, in the light of (\ref{e:expectation}), and the fact that the conditional variance of $\frac{1}{|N_{i_t}({\bar \bx_t})|}$ is not larger than its conditional mean, we can use, e.g., \cite{kt08} to conclude that, with probability at least $1-\delta$,
\begin{align*}
\sum_{t=1}^T \frac{1}{|N_{i_t}({\bar \bx_t})|} \leq \frac{2Tc}{n}\,\E\left[m(X)\right]  + 12\,\log\left(\frac{\log T}{\delta}\right)\,,
\end{align*}
as claimed. $\hfill\Box$

\renewcommand{\S}{S}
\newcommand{\bbbb}{\bm{b}}
\newcommand{\bbt}{\bbbb^t}
\newcommand{\bg}{\bm{g}}
\newcommand{\gt}{\bg^t}
\newcommand{\s}{s}
\newcommand{\bt}{\bw}
\newcommand{\St}{\S^t}
\newcommand{\bto}{\bt^*}
\newcommand{\btt}{\bt^t}
\newcommand{\ttn}{\bt^{t+1}}
\newcommand{\ip}[2]{\langle #1, #2 \rangle}
\newtheorem{definition}{Definition}
\renewcommand{\ss}{L}
\renewcommand{\sc}{\alpha}
\newcommand{\So}{\S^*}
\newcommand{\so}{s^*}
\newcommand{\aps}{\alpha^\text{sp}}
\newcommand{\scb}{\text{sCB}}
\newcommand{\mycb}{\text{CB}}

\section{Extending CAB to Sparse User Models}
\label{sec:sparse}
In this section, we give details on how CAB can be modified to work when user models (i.e. the vectors $\bu_i, i = 1, \ldots ,n$) are $s$-sparse i.e. $\|\bu_i\|_0 \leq s$ for $s \ll d$. We will denote $S_i = supp(\bu_i)$ to be the support of the vector for user $i$. We will assume for the sake of simplicity that $|S_i| \leq \so$ for all $i$. We will also make the standard assumption that non-zero coordinates of the vectors do not take vanishing values. More formally, we will assume that for some $\pi > 0$, for all $i$, for all $j$, either $\bu_i[j] = 0$ or else $|\bu_i[j]| > \pi$. Note that different users can have different supports, but all of them must be $\so$-sparse.

Sparse user models arise when the user and item vectors are extremely high dimensional and not all features are useful in encoding the preference patterns of every user. Rather, every user chooses a (possibly different) set of features that best encode its preferences. Sparse models are also extremely popular in resource constrained settings where dense models are too expensive to store or too slow to predict with.

In such cases, performing least squares regression to obtain the proxy vectors is not only expected to give poor results, but also requires the much larger number of trials $\scO(d)$ per user $i$ to effectively estimate $\bu_i$, which can be prohibitive since users typically interact very sparsely with recommendation systems.

To make our exposition easier, we introduce some handy notation. Let $T_i(t) = \sum_{\tau=1}^{t-1}\{i_t = i\}$ denote the number of times user $i$ has been served till time $t$. Also let $X_{i,t} \in \R^{T_i(t) \times d}$ denote the matrix of all item context vectors that have been served to user $i$ till time $t$ and $\epsilon_{i,t} \in \R^{T_i(t)}$ denote the vector of sub-Gaussian error values that were introduced into the payoffs user $i$ offered in the past. Also, for any set $S \subset [d]$, we let $X_{i,t}^S$ denote the submatrix of $X_{i,t}$ that contains only those columns that are present in the set $S$ and the rest zeroed out. For any vector $\bu$, the notation $\bu^S$ will denote the vector with coordinates in $S$ retained and the rest zeroed out. However, we will abuse this notation while using it in context of the correlation matrix. We will denote using $M_{i,t-1}^{S} = (X_{i,t}^S)^\top X_{i,t}^S$ the correlation matrix formed using item context vectors restricted to the coordinates in $S$.

To ameliorate the challenges in high dimensional settings with sparse user models, we present the spCAB algorithm which adapts to extremely high dimensional features. The spCAB algorithm is identical to the CAB algorithm (Algorithm~\ref{alg:tcab} except in two critical respects
\begin{enumerate}
	\item In step 4, instead of solving the least squares problem to identify the next proxy for the users, a sparse recovery technique is used, given in \eqref{eq:sprec} below.
	\item We use a different notion of confidence bounds $\scb_{i,t}(\bx) = \aps(t)\,\sqrt{\bx^\top (M_{i,t-1}^{\hat S_{i,t-1}})^{-1} \bx}$, where $\hat S_{i,t-1} = supp(\bw_{i,t-1})$ is the support of the current estimate of the model for user $i$, and we set $\aps(t) \leq \sqrt{s\log T}$.
\end{enumerate}

\begin{align}
\bw_{i,t-1} = \min_{\norm{\bw}_0 \leq s'} f_{i,t}(\bw) =: \sum_{t: i_t = i}(\bw^\top\bar\bx_t - y_t)^2 + \|\bw\|_2^2
\label{eq:sprec}
\end{align}

Sparse recovery has a rich history in signal processing and learning domains with a tremendous amount of progress in recent years. There exist a plethora of methods, including relaxation techniques, iterative hard thresholding techniques, pursuit techniques and fully corrective techniques, to solve the problem.

For our purposes, fully corrective methods such as CoSaMP \cite{NeedellT08} and Subspace Pursuit \cite{DaiM09} would be very convenient. These offer a linear rate of convergence whenever requisite properties, mentioned below, are satisfied. Algorithm~\ref{algo:tstage} gives a general outline of these methods for general sparse recovery with an objective function $f$. We will only be required to consider the case when $f$ is the ridge-regression function induced in the linear bandit problem as mentioned in \eqref{eq:sprec}.

\begin{algorithm}[t]
\caption{Two-stage Hard-thresholding}
  \begin{algorithmic}[1]
    \STATE {\bf Input}: function $f$ with gradient oracle, sparsity level $\s$, sparsity expansion level $\ell$
    \STATE $\bt^1=0$, $t=1$
    \WHILE{\emph{not converged}}
    \STATE $\gt=\nabla_{\bt}f(\btt)$, $\St=supp(\btt)$
    \STATE $Z^t=\St\cup (\text{largest } \ell \text{ elements of }|\gt_{\overline{\St}}|)$
    \STATE $\bbt=\arg\min_{\bm{\beta}, supp(\beta)\subseteq Z^t}f(\bm{\beta})$ \hfill // fully corrective step%
    \STATE $\widetilde{\bt}^t=P_\s(\bbt)$
    \STATE $\ttn=\arg\min_{\bt, supp(\bt)\subseteq supp(\widetilde{\bt}^t)}f(\bt)$ \hfill // fully corrective step
    \STATE $t=t+1$
	\ENDWHILE 
    \STATE {\bf Output}: $\btt$
  \end{algorithmic}\label{algo:tstage}
\end{algorithm}

Two properties that would be crucial to analyzing these sparse recovery methods are those of restricted strong convexity and restricted strong smoothness, outlined below.
\begin{definition}[RSC Property]
\label{defn:rsc}
A differentiable function $f:\R^p\rightarrow \R$ is said to satisfy restricted strong convexity (RSC) at sparsity level $s = \s_1+\s_2$ with strong convexity constraint $\sc_{s}$ if the following holds for all $\bt_1, \bt_2$ s.t. $\|\bt_1\|_0\leq \s_1$ and $\|\bt_2\|_0\leq \s_2$: 
$$f(\bt_1)-f(\bt_2)\geq \ip{\bt_1-\bt_2}{\nabla_{\bt} f(\bt_2)}+\frac{\sc_{s}}{2}\|\bt_1-\bt_2\|_2^2.$$
\end{definition}
\begin{definition}[RSS Property]
\label{defn:rss}
A differentiable function $f:\R^p\rightarrow \R$ is said to satisfy restricted strong smoothness (RSS) at sparsity level $s = \s_1+\s_2$ with strong convexity constraint $\ss_{s}$ if the following holds for all $\bt_1, \bt_2$ s.t. $\|\bt_1\|_0\leq \s_1$ and $\|\bt_2\|_0\leq \s_2$:
$$f(\bt_1)-f(\bt_2)\leq \ip{\bt_1-\bt_2}{\nabla_{\bt} f(\bt_2)}+\frac{\ss_{s}}{2}\|\bt_1-\bt_2\|_2^2.$$ 
\end{definition}

\subsection{Regret Analysis for spCAB}
We will sketch a regret bound proof for spCAB by proving counterparts to Lemmata~\ref{l:anc1} and \ref{l:anc3} in the sparse user model case. Lemma~\ref{l:anc2} will not require any modifications. First of all we invoke sparse recovery guarantees \citep[Theorems 3 and 4]{JainTK2014} and standard martingale arguments to show the following result for two-stage fully corrective methods when applied to the user proxy estimation problem.
\begin{theorem}
\label{thm:tstage}
Suppose the objective function $f_{i,t}$ satisfies RSC and RSS parameters given by $\sc_{2\s+\so}(f_{i,t})=\sc$ and $\ss_{2\s+\ell}(f_{i,t})=\ss$ respectively. Suppose Algorithm~\ref{algo:tstage} is invoked with $f_{i,t}$, $\ell \geq \so$ and $\s\geq 4\frac{\ss^2}{\sc^2}\ell+\so-\ell\geq 4\frac{\ss^2}{\sc^2}\so$. Then, the $\tau$-th iterate of Algorithm~\ref{algo:tstage}, for $\tau = O(\frac{\ss}{\sc}\cdot\log(\frac{1}{\epsilon}))$ satisfies, with probability at least $1 - \delta$: $\|\bt^\tau-\bu_i\|_2 \leq O\Bigl(\sqrt{\frac{\so \log \frac{d}{\delta}}{T_i(t)}} + \sqrt\epsilon\Bigl)$.
\end{theorem}

It is easy to see that if we run Algorithm~\ref{algo:tstage} for longer than $\Omega(\frac{1}{\pi^2})$ iterations, as well as if $T_i(t) \geq \Omega\Bigl(\frac{\so\log dT}{\pi^2}\Bigl)$, then, we will have, with very high probability,
\[
\|\bt^\tau-\bu_i\|_2 \leq \pi/2,
\]
which can be easily seen to guarantee that $\hat S_{i,t} := supp(\bw_{i,t-1}) = supp(\bt^\tau) = supp(\bu_i) = S_i$, since we set $\bw_{i,t-1} = \bt^\tau$. Now, notice that fully corrective methods always solve the least squares problem over their current support. This means that if we denote $X = X_{i,t}$, then we have
\[
\bt^\tau = ((X^{S_i})^\top X^{S_i} + I)^{-1}(X^{S_i})^\top(X\bu_i + \epsilon_{i,t}) = ((X^{S_i})^\top X^{S_i} + I)^{-1}(X^{S_i})^\top(X^{S_i}\bu_i^{S_i} + \epsilon_{i,t})
\]
Using a proof technique identical to the one used for proving \citep[Theorem 2]{abbasi2011improved} and applying restricted strong convexity, we can show that, for any $\bx$
\[
|\bw_{i,t-1}^\top\bx - \bu_i^\top\bx| \leq \aps(t)\cdot\sqrt{\bx^\top (M_{i,t-1}^{\hat S_{i,t}})^{-1}\bx},
\]
where $\aps(t) \leq O(\sqrt{s\log T/\delta})$. Notice the stark improvement in the behavior of the exploration parameter in the sparse model case. For the dense model, we had to set $\alpha(t) \sim \sqrt{d \log t}$ and in high dimensional settings, we have $\aps(t) \ll \alpha(t)$. Note also that this establishes the counterpart to Lemma~\ref{l:anc3} in the sparse model case.

To complete the regret bound proof for the sparse case, we need to now complete three tasks:
\begin{enumerate}
	\item Establish Lemma~\ref{l:anc1};
	\item Satisfy the RSC, RSS conditions;
	\item Ensure that $T_i(t) \geq \Omega\Bigl(\frac{\so\log dT}{\pi^2}\Bigl)$.
\end{enumerate}
The third task is the simplest -- it is easy to see that using standard results on Bernoulli variables, after $T_0 = 2n\Bigl(\frac{\so\log dT}{\pi^2} + 2\log\frac{nT}{\delta}\Bigl)$, with probability at least $1-\delta$, $T_i(t) \geq \Omega\Bigl(\frac{\so\log dT}{\pi^2}\Bigl)$ for all $i \in \scU$ and all $t > T_0$.

It turns out that the first and the second tasks are actually identical. Recall from the proof of Lemma~\ref{l:anc1} that all we need is an upper bound on the quantity $\scb_{i,t}(\bx)$. While bounding $\mycb$, this was equivalent to establishing an upper bound on $\sqrt{\bx^\top M_{i,t}^{-1}\bx}$ which was itself equivalent to a lower bound on the eigenvalues of $M_{i,t}$. In this case, we need to show an upper bound on
\[
\sqrt{\bx^\top (M^{\hat S_{i,t}}_{i,t})^{-1}\bx} = \sqrt{(\bx^{\hat S_{i,t}})^\top(M^{\hat S_{i,t}}_{i,t})^{-1}\bx^{\hat S_{i,t}}},
\]
since $M_{i,t}^S = (X_{i,t}^S)^\top X_{i,t}^S$. This is equivalent to showing a lower bound on the \emph{restricted eigenvalues} of $M_{i,t}$. More specifically, we will want to establish a lower bound on the following quantity:
\[
\min_{S \subset [d], |S| = \so}\lambda_{\min}(M^S_{i,t})
\]
It is easy to see that this is equivalent to demonstrating RSC/RSS properties of the objective function $f_{i,t}$. Corresponding to the restricted eigenvalue requirement, we propose a counterpart of the hardness coefficient $\text{spHD}\Bigl(\{i_t,C_t\}_{t=1}^T, \eta\Bigl)$, wherein we wish an upper bound on the time before all possible correlation matrices of all users have their $\so$-restricted eigenvalues bounded below by $\eta$. For our results, we would require a bound on $\text{spHD}\Bigl(\{i_t,C_t\}_{t=1}^T, \frac{16(\aps(T))^2}{\gamma^2}\Bigl)$. Fortunately, the technique of \citet{glz14} of using Freedman-style inequalities \cite{Tropp2011} which was used to prove the bound for Lemma~\ref{l:anc1} can still be harnessed to give
\[
\text{sHD}\Bigl(\{i_t,C_t\}_{t=1}^T, \eta\Bigl) \leq O\Bigl(\frac{n\so\eta}{\lambda^2}\log\frac{Tnd}{\delta\so}\Bigl)
\]
Using the above, we can show the following regret bound for spCAB in the sparse user model setting.
\begin{theorem}
\label{c:sp}
If spCAB is executed on a bandit clustering setting with $\so$-sparse user models satisfying the requisite properties mentioned above, then with probability at least $1-\delta$, the regret of spCAB satisfies
\[
\sum_{t=1}^T r_t \leq R^\text{sp} + {\tilde \scO}\left(\sqrt{\so\,d\,Tc\left(\E[m(X)] \right)} \right)~,
\]
where the ${\tilde \scO}$-notation hides logarithmic factors in $\frac{TNd}{\delta}$, and $R$ is of the form
\[
R^\text{sp} = \frac{c\,n^2\,(\so)^2\sqrt\so}{\lambda^2\,\gamma^2}\,\log^{2.5}\left(\frac{Tnd}{\delta\so}\right).
\]
\end{theorem}
Notice the drastic reduction in the dependence of $d$ in the regret bound. spCAB enjoys a regret bound that depends weakly on the ambient dimension of the problem setting as it has only $\sqrt{d}$ and $\log d$ dependence on $d$. We also notice that the bound can be further improved if the item contexts are low dimensional as well.

Suppose that the item contexts $\bx_{t,k}$ are sampled from a distribution that has support only over a low $r$-dimensional space. Then, it can be easily shown that spCAB, without any modifications, offers the following regret bound.
\[
\sum_{t=1}^T r_t \leq \frac{c\,n^2\,(\so)^2\sqrt\so}{\lambda^2\,\gamma^2}\,\log^{2.5}\left(\frac{Tnd}{\delta\so}\right) + {\tilde \scO}\left(\sqrt{\so\,r\,Tc\left(\E[m(X)] \right)} \right)~.
\]
Note that in this regret bound, the dependence on $d$ is only in logarithmic terms.

\end{document}